\documentclass{article}

\usepackage[nonatbib,final]{neurips_data_2021}

\usepackage[utf8]{inputenc} 
\usepackage[T1]{fontenc}    
\usepackage{hyperref}       
\usepackage{url}            
\usepackage{booktabs}       
\usepackage{amsfonts}       
\usepackage{nicefrac}       
\usepackage{microtype}      
\usepackage{xcolor}         
\usepackage{subcaption}
\usepackage{float}
\newcommand{\comment}[1]{}
\usepackage{tabularx}
\newcolumntype{Y}{>{\centering\arraybackslash}X}
\usepackage{wrapfig}
\usepackage{mdframed}
\usepackage[pdftex]{graphicx}
\usepackage{pgfplots}
\usepackage{adjustbox}
\usepackage{amsmath}
\usepackage{amssymb}
\usepackage{bbm}
\usepackage{bm}
\usepackage{multirow}
\usepackage{textpos}
\usepackage{subcaption}
\usepackage{xcolor}
\usepackage{setspace}
\usepackage{listings}
\usepackage[frozencache=true,cachedir=.]{minted}

\newcommand{\dssectionheader}[1]{%
   \noindent\framebox[\columnwidth]{%
      {\fontfamily{phv}\selectfont \textbf{\textcolor{blue}{#1}}}
   }
}

\newcommand{\dsquestion}[1]{%
   {\noindent \scriptsize {\fontfamily{phv}\selectfont \textcolor{blue}{\textbf{#1}}}}
}

\newcommand{\dsquestionex}[2]{%
   {\noindent \scriptsize {\fontfamily{phv}\selectfont \textcolor{blue}{\textbf{#1} #2}}}
}

\newcommand{\dsanswer}[1]{%
   {\noindent \footnotesize {#1} \medskip}
}

\title{The Multi-Agent Behavior Dataset: Mouse Dyadic Social Interactions}

\author{%
Jennifer J. Sun\\
Caltech \\
\And
Tomomi Karigo\\
Caltech \\
\And
Dipam Chakraborty\\
AICrowd Research \\
\And
Sharada P. Mohanty\\
AICrowd Research \\
\And
Benjamin Wild \\
Freie Universität Berlin \\
\And
Quan Sun \\
OPPO Research Institute \\
\And
Chen Chen\\
OPPO Research Institute \\
\And
David J. Anderson\\
Caltech \\
\And
Pietro Perona\\
Caltech \\
\And
Yisong Yue\\
Caltech \\
\And
Ann Kennedy\\
Northwestern University\\
\texttt{ann.kennedy@northwestern.edu} \\
\And
{\small Dataset Website: \url{https://sites.google.com/view/computational-behavior/our-datasets/calms21-dataset}} 
}

\begin{document}

\maketitle

\begin{abstract}

Multi-agent behavior modeling aims to understand the interactions that occur between agents. We present a multi-agent dataset from behavioral neuroscience, the Caltech Mouse Social Interactions (CalMS21) Dataset. Our dataset consists of trajectory data of social interactions, recorded from videos of freely behaving mice in a standard resident-intruder assay. 
To help accelerate behavioral studies, the CalMS21 dataset provides benchmarks to evaluate the performance of automated behavior classification methods in three settings: (1) for training on large behavioral datasets all annotated by a single annotator, (2) for style transfer to learn inter-annotator differences in behavior definitions, and (3) for learning of new behaviors of interest given limited training data. The dataset consists of 6 million frames of unlabeled tracked poses of interacting mice, as well as over 1 million frames with tracked poses and corresponding frame-level behavior annotations. The challenge of our dataset is to be able to classify behaviors accurately using both labeled and unlabeled tracking data, as well as being able to generalize to new settings. 

\end{abstract}


\section{Introduction}

The behavior of intelligent agents is often shaped by interactions with other agents and the environment. As a result, models of multi-agent behavior are of interest in diverse domains, including neuroscience~\cite{segalin2020mouse}, video games~\cite{hofmann2019minecraft}, sports analytics~\cite{zhan2019learning}, and autonomous vehicles~\cite{chang2019argoverse}. Here, we study multi-agent animal behavior from neuroscience and introduce a dataset to benchmark behavior model performance.  

Traditionally, the study of animal behavior relied on the manual, frame-by-frame annotation of behavioral videos by trained human experts. This is a costly and time-consuming process, and cannot easily be crowdsourced due to the training required to identify many behaviors accurately. Automated behavior classification is a popular emerging tool~\cite{kabra2013jaaba,anderson2014toward, eyjolfsdottir2016learning, nilsson2020simple, segalin2020mouse}, as it promises to reduce human annotation effort, and opens the field to more high-throughput screening of animal behaviors. However, there are few large-scale publicly available datasets for training and benchmarking social behavior classification, and the behaviors annotated in those datasets may not match the set of behaviors a particular researcher wants to study. Collecting and labeling enough training data to reliably identify a behavior of interest remains a major bottleneck in the application of automated analyses to behavioral datasets.

\begin{wrapfigure}{r}{0.54\linewidth}
    \centering
  \includegraphics[width=\linewidth]{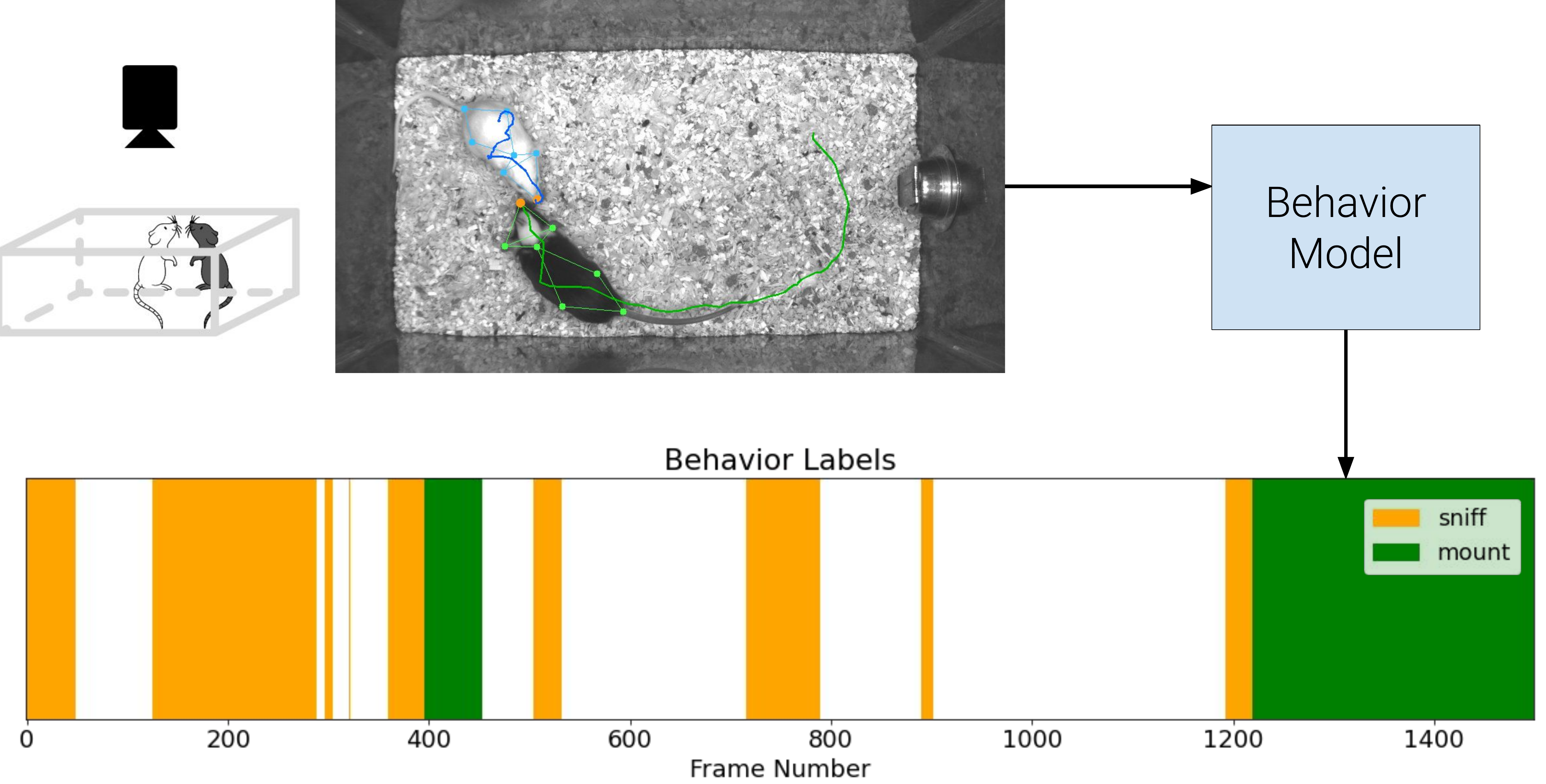}
  \caption{{\bf Overview of behavior classification.} A typical behavior study starts with extraction of tracking data from videos. We show 7 keypoints for each mouse, and draw the trajectory of the nose keypoint. The goal of the model is to classify each frame (30Hz) to one of the behaviors of interest from domain experts. }
  \label{fig:introduction}
\end{wrapfigure}

We present a dataset of behavior annotations and tracked poses from pairs of socially interacting mice, the \textbf{Cal}tech \textbf{M}ouse \textbf{S}ocial Interactions 2021 (CalMS21) Dataset, with the goal of advancing the state-of-the-art in behavior classification. From top-view recorded videos of mouse interactions, we detect seven keypoints for each mouse in each frame using Mouse Action Recognition System (MARS)~\cite{segalin2020mouse}. Accompanying the pose data, we introduce three tasks pertaining to the classification of frame-level social behavior exhibited by the mice, with frame-by-frame manual annotations of the behaviors of interest (Figure~\ref{fig:introduction}), and additionally release video data for a subset of the tasks. Finally, we release a large dataset of tracked poses without behavior annotations, that can be used to study unsupervised learning methods. 

As part of the initial benchmarking of CalMS21, we both evaluated standard baseline methods as well as solicited novel methods by having CalMS21 as part of the Multi-Agent Behavior (MABe) Challenge 2021 hosted at CVPR 2021. To test model generalization, our dataset contains splits annotated by different annotators and for different behaviors. 

In addition to providing a performance benchmark for multi-agent behavior classification, our dataset is suitable for studying several research questions, including: How do we train models that transfer well to new conditions (annotators and behavior labels)? How do we train accurate models to identify rare behaviors? How can we best leverage unlabeled data for behavior classification?


\section{Related Work}

\textbf{Behavior Classification.} Automated behavior classification tools are becoming increasingly adopted in neuroscience~\cite{segalin2020mouse,eyjolfsdottir2014detecting,kabra2013jaaba,anderson2014toward,nilsson2020simple}. These automated classifiers generally consists of the following steps: pose estimation, feature computation, and behavior classification. Our dataset provides the output from our mouse pose tracker, MARS~\cite{segalin2020mouse}, to allow participants in our dataset challenge to focus on developing methods for the latter steps of feature computation and behavior classification. We will therefore first focus our exploration of related works on these two topics, specifically within the domain of neuroscience, then discuss how our work connects to the related field of human action recognition.

Existing behavior classification methods are typically trained using tracked poses or hand-designed features in a fully-supervised fashion with human-annotated behaviors~\cite{hong2015automated,segalin2020mouse,burgos2012social,eyjolfsdottir2014detecting,nilsson2020simple}. Pose representations used for behavior classification can take the form of anatomically defined keypoints~\cite{segalin2020mouse,nilsson2020simple}, fit shapes such as ellipses~\cite{hong2015automated,dankert2009automated,ohayon2013automated}, or simply a point reflecting the location of an animal's centroid~\cite{noldus2001ethovision,perez2014idtracker,wild2021social}. Features extracted from poses may reflect values such as animal velocity and acceleration, distances between pairs of body parts or animals, distances to objects or parts of the arena, and angles or angular velocities of keypoint-defined joints. To bypass the effort-intensive step of hand-designing pose features, self-supervised methods for feature extraction have been explored~\cite{sun2020task}. Computational approaches to behavior analysis in neuroscience have been recently reviewed in ~\cite{pereira2020quantifying, datta2019computational, egnor2016computational, anderson2014toward}.

Relating to behavior classification and works in behavioral neuroscience, there is also the field of human action recognition (reviewed in \cite{wu2017recent,zhang2019comprehensive}). We compare this area to our work in terms of models and data. Many works in action recognition are trained end-to-end from image or video data~\cite{luvizon20182d,simonyan2014two,tran2015learning,carreira2017quo}, and the sub-area that is most related to our dataset is pose/skeleton-based action recognition~\cite{choutas2018potion,liu2020disentangling,sun2020view}, where model input is also pose data. However, one difference is that these models often aim to predict one action label per video, since in many datasets, the labels are annotated at the video or clip level~\cite{carreira2017quo,zhang2013actemes,shahroudy2016ntu}. More closely related to our dataset is the works on temporal segmentation~\cite{kukleva2019unsupervised,stein2013combining,kuehne2014language}, where one action label is predicted per frame in long videos. These works are based on human activities, often goal-directed in a specific context, such as cooking in the kitchen. 
We would like to note a few unique aspects animal behavior. In contrast to many human action recognition datasets, naturalistic animal behavior often requires domain expertise to annotate, making it more difficult to obtain. Additionally, most applications of animal behavior recognition are in scientific studies in a laboratory context, meaning that the environment is under close experimenter control.

\textbf{Unsupervised Learning for Behavior.} As an alternative to supervised behavior classification, several groups have used unsupervised methods to identify actions from videos or pose estimates of freely behaving animals~\cite{berman2014mapping,klibaite2017unsupervised,wiltschko2015mapping,vogelstein2014discovery,luxem2020identifying,hsu2020b,marques2018structure} (also reviewed in ~\cite{datta2019computational, pereira2020quantifying}). In most unsupervised approaches, videos of interacting animals are first processed to remove behavior-irrelevant features such as the absolute location of the animal; this may be done by registering the animal to a template or extract a pose estimate. Features extracted from the processed videos or poses are then clustered into groups, often using a model that takes into account the temporal structure of animal trajectories, such as a set of wavelet transforms~\cite{berman2014mapping}, an autoregressive hidden Markov model~\cite{wiltschko2015mapping}, or a recurrent neural network~\cite{luxem2020identifying}. Behavior clusters produced from unsupervised analysis have been shown to be sufficiently sensitive to distinguish between animals of different species and strains~\cite{hernandez2020framework,luxem2020identifying}, and to detect effects of pharmacological perturbations~\cite{wiltschko2020revealing}. Clusters identified in unsupervised analysis can often be related back to human-defined behaviors via post-hoc labeling ~\cite{wiltschko2015mapping,berman2014mapping,vogelstein2014discovery}, suggesting that cluster identities could serve as a low-dimensional input to a supervised behavior classifier.

\textbf{Related Datasets.} The CalMS21 dataset provides a benchmark to evaluate the performance of behavior analysis models. Related animal social behavior datasets include CRIM13~\cite{burgos2012social} and Fly vs. Fly~\cite{eyjolfsdottir2014detecting}, which focus on supervised behavior classification. In comparison to existing datasets, CalMS21 enables evaluation in multiple settings, such as for annotation style transfer and for learning new behaviors. The trajectory data provided by the MARS tracker~\cite{segalin2020mouse} (seven keypoints per mouse) in our dataset also provides a richer description of the agents compared to single keypoints (CRIM13). Additionally, CalMS21 is a good testbed for unsupervised and self-supervised models, given its inclusion of a large (6 million frame) unlabeled dataset. 

While our task is behavior classification, we would like note that there are also a few datasets focusing on the related task of multi-animal tracking~\cite{pedersen20203d,graving2019deepposekit}. Multi-animal tracking can be difficult due to occlusion and identity tracking over long timescales. In our work, we used the output of the MARS tracker~\cite{segalin2020mouse}, which also includes a multi-animal tracking dataset on the two mice to evaluate pose tracking performance; we bypass the problem of identity tracking by using animals of differing coat colors. Improved methods to more accurately track multi-animal data is another direction that can help quantify animal behavior.

\begin{figure*}[t]
    \centering
  \includegraphics[width=\linewidth]{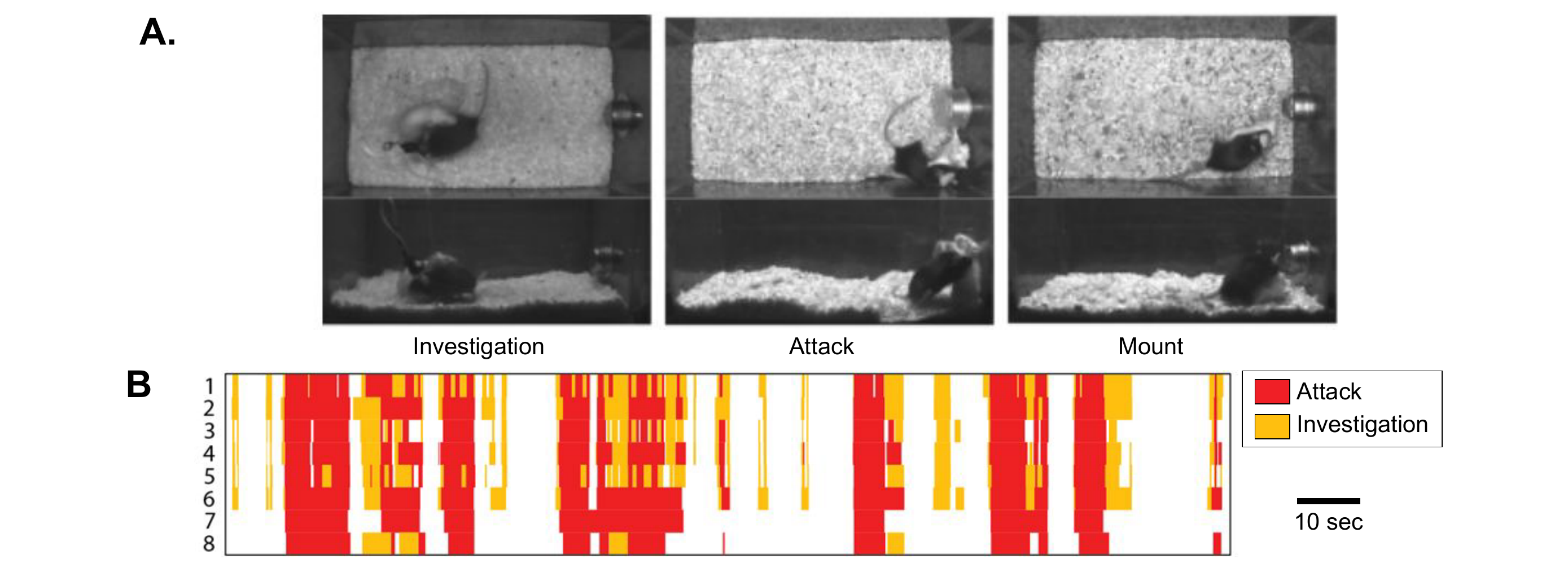}
  \caption{{\bf Behavior classes and annotator variability.} A. Example frames showing some behaviors of interest. B. Domain expert variability in behavior annotation, reproduced with permission from ~\cite{segalin2020mouse}. Each row shows annotations from a different domain expert annotating the same video data.}
  \label{fig:samples}
\end{figure*}

Other datasets studying multi-agent behavior include those from autonomous driving~\cite{chang2019argoverse,sun2020scalability}, sports analytics~\cite{yue2014learning,decroos2018automatic}, and video games~\cite{samvelyan2019starcraft,guss2019minerl}. Generally, the autonomous vehicle datasets focus on tracking and forecasting, whereas trajectory data is already provided in CalMS21, and our focus is on behavior classification. Sports analytics datasets also often involves forecasting and learning player strategies. Video game datasets have perfect tracking and generally focus on reinforcement learning or imitation learning of agents in the simulated environment. While the trajectories in CalMS21 can be used for imitation learning of mouse behavior, our dataset also consist of expert human annotations of behaviors of interest used in scientific experiments. As a result, CalMS21 can be used to benchmark supervised or unsupervised behavior models against expert human annotations of behavior.

\section{Dataset Design}

\begin{wrapfigure}{r}{0.5\linewidth}
    \centering
  \includegraphics[width=0.75\linewidth]{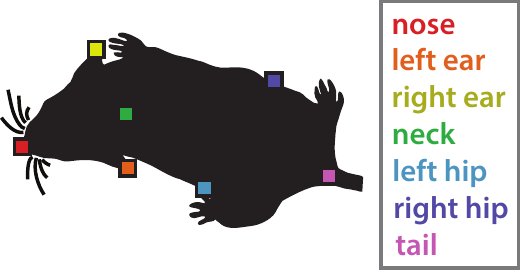}
  \caption{{\bf Pose keypoint definitions.} Illustration of the seven anatomically defined keypoints tracked on the body of each animal. Pose estimation is performed using MARS~\cite{segalin2020mouse}.}
  \label{fig:keypoint_ids}
\end{wrapfigure}

The CalMS21 dataset is designed for studying behavior classification, where the goal is to assign frame-wise labels of animal behavior to temporal pose tracking data. The tracking data is a top-view pose estimate of a pair of interacting laboratory mice, produced from raw 30Hz videos using MARS~\cite{segalin2020mouse}, and reflecting the location of the nose, ears, neck, hips, and tail base of each animal (Figure \ref{fig:keypoint_ids}).

We define three behavior classification tasks on our dataset. In Task 1 (Section~\ref{sec:task_1}), we evaluate the ability of models to classify three social behaviors of interest (attack, mount, and close investigation) given a large training set of annotated videos; sample frames of the three behaviors are shown in Figure~\ref{fig:samples}A. In Task 2 (Section~\ref{sec:task_2}), models must be adjusted to reproduce new annotation styles for the behaviors studied in Task 1: Figure~\ref{fig:samples}B demonstrates the annotator variability that can exist for the same videos with the same labels. Finally, in Task 3 (Section~\ref{sec:task_3}), models must be trained to classify new social behaviors of interest given limited training data.

\begin{figure*}[!t]
   \centering
    \begin{subfigure}{\textwidth}
   {\includegraphics[width=\linewidth]{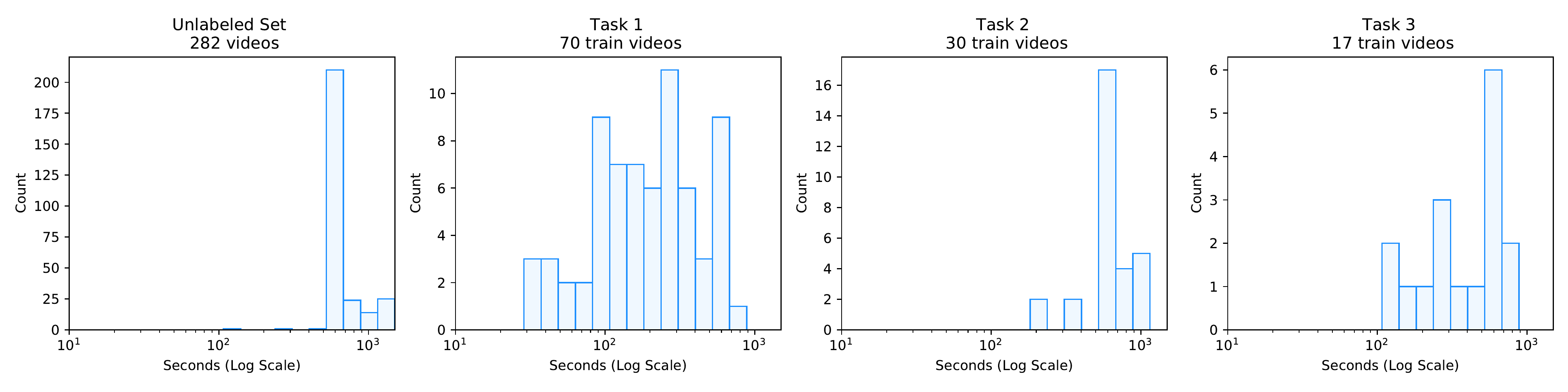}}
   \caption{Video Length Distribution}
   \end{subfigure}
    \begin{subfigure}{\textwidth}      
   {\includegraphics[width=1.05\linewidth]{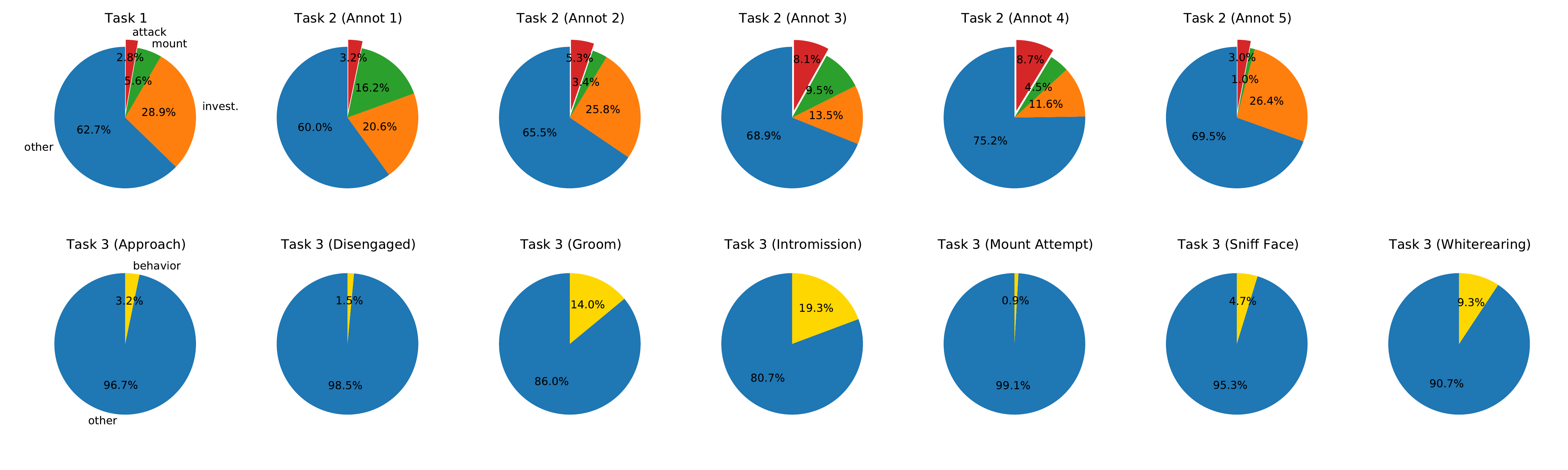}}  
   \caption{Percentage of Annotated Behaviors}
    \end{subfigure}      
   \caption{{\bf Available data for each task in our challenge.} Our dataset consists of a large set of unlabeled videos alongside a set of annotated videos from one annotator. Annot 1, 2, 3, 4, 5 are different domain experts, whose annotations for attack, mount, and investigation are used in Task 2. Bottom row shows new behaviors used in Task 3.}
   \label{fig:data_description}
\end{figure*}

In Tasks 1 and 2, each frame is assigned one label (including "other" when no social behavior is shown), therefore these tasks can be handled as multi-class classification problems. In Task 3, we provide separate training sets for each of seven novel behaviors of interest, where in each training set only a single behavior has been annotated. For this task, model performance is evaluated for each behavior separately: therefore, Task 3 should be treated as a set of binary classification problems. Behaviors are temporal by nature, and often cannot be accurately identified from the poses of animals in a single frame of video. Thus, all three tasks can be seen as time series prediction problems or sequence-to-sequence learning, where the time-evolving trajectories of 28-dimensional animal pose data (7 keypoints x 2 mice x 2 dimensions) must be mapped to a behavior label for each frame. Tasks 2 and 3 are also examples of few-shot learning problems, and would benefit from creative forms of data augmentation, task-efficient feature extraction, or unsupervised clustering to stretch the utility of the small training sets provided.

To encourage the combination of supervised and unsupervised methods, we provide a large set of unlabeled videos (around 6 million frames) that can be used for feature learning or clustering in any task (Figure~\ref{fig:data_description}).

\subsection{Task 1: Classical Classification}~\label{sec:task_1}
Task 1 is a standard sequential classification task: given a large training set comprised of pose trajectories and frame-level annotations of freely interacting mice, we would like to produce a model to predict frame-level annotations from pose trajectories on a separate test set of videos. There are 70 videos in the public training set, all of which have been annotated for three social behaviors of interest: close investigation, attack, and mount (described in more detail in the appendix). The goal is for the model to reproduce the behavior annotation style from the training set.

Sequential classification has been widely studied, existing works use models such as recurrent neural networks~\cite{chung2014empirical}, temporal convolutional networks~\cite{lea2017temporal}, and random forests with hand-designed input features~\cite{segalin2020mouse}. Input features to the model can also be learned with self-supervision~\cite{sun2020task,chen2020simple,liu2020self}, which can improve classification performance using the unlabeled portion of the dataset.

In addition to pose data, we also release all source videos for Task 1, to facilitate development of methods that require visual data.

\subsection{Task 2: Annotation Style Transfer}~\label{sec:task_2}
In general, when annotating the same videos for the same behaviors, there exists variability across annotators (Figure~\ref{fig:samples}B). As a result, models that are trained for one annotator may not generalize well to other annotators. Given a small amount of data from several annotators, we would like to study how well a model can be trained to reproduce each individual's annotation style. Such an "annotation style transfer" method could help us better understand differences in the way behaviors are defined across annotators and labs, increasing the reproducibility of experimental results. A better understanding of different annotation styles could also enable crowdsourced labels from non-experts to be transferred to the style of expert labels.

In this sequential classification task, we provide six ~10-minute training videos for each of five annotators unseen in Task 1, and evaluate the ability of models to produce annotations in each annotator's style. All annotators are trained annotators from the David Anderson Lab, and have between several months to several years of prior annotation experience. The behaviors in the training datasets are the same as Task 1. In addition to the annotator-specific videos, competitors have access to a) the large collection of task 1 videos, that have been annotated for the same behaviors but in a different style, and b) the pool of unannotated videos, which could be used for unsupervised clustering or feature learning.

This task is suitable for studying techniques from transfer learning~\cite{tan2018survey} and domain adaptation~\cite{wang2018deep}. We have a source domain with labels from task 1, which needs to be transferred to each annotator in task 2 with comparatively fewer labels. Potential directions include learning a common set of data-efficient features for both tasks~\cite{sun2020task}, and knowledge transfer from a teacher to a student network~\cite{ahn2019variational}.

\subsection{Task 3: New Behaviors}~\label{sec:task_3}
It is often the case that different researchers will want to study different behaviors in the same experimental setting. The goal of Task 3 is to help benchmark general approaches for training new behavior classifiers given a small amount of data. This task contains annotations on seven behaviors not labeled in Tasks 1 \& 2, where some behaviors are very rare (Figure~\ref{fig:data_description}).

As for the previous two tasks, we provide a training set of videos in which behaviors have been annotated on a frame-by-frame basis, and evaluate the ability of models to produce frame-wise behavior classifications on a held-out test set. We expect that the large unlabeled video dataset (Figure~\ref{fig:data_description}) will help improve performance on this task, by enabling unsupervised feature extraction or clustering of animal pose representations prior to classifier training. 

Since each new behavior has a small amount of data, few-show learning techniques~\cite{wang2020generalizing} can be helpful for this task. The data from Task 1 and the unlabeled set could also be used to set up multi-task learning~\cite{zhang2017survey}, and for knowledge transfer~\cite{ahn2019variational}.


\section{Benchmarks on CalMS21}~\label{sec:benchmarks}

We develop an initial benchmark on CalMS21 based on standard baseline methods for sequence classification. To demonstrate the utility of the unlabeled data, we also used these sequences to train a representation learning framework (task programming~\cite{sun2020task}) and added the learned trajectory features to our baseline models. Additionally, we presented CalMS21 at the MABe Challenge hosted at CVPR 2021, and we include results on the top performing methods for each of the tasks. 

Our evaluation metrics are based on class-averaged F1 score and Mean Average Precision (more details in the appendix). Unless otherwise stated, the class with the highest predicted probability in each frame was used to compute F1 score.

\begin{figure*}
    \centering
    \includegraphics[width=\linewidth]{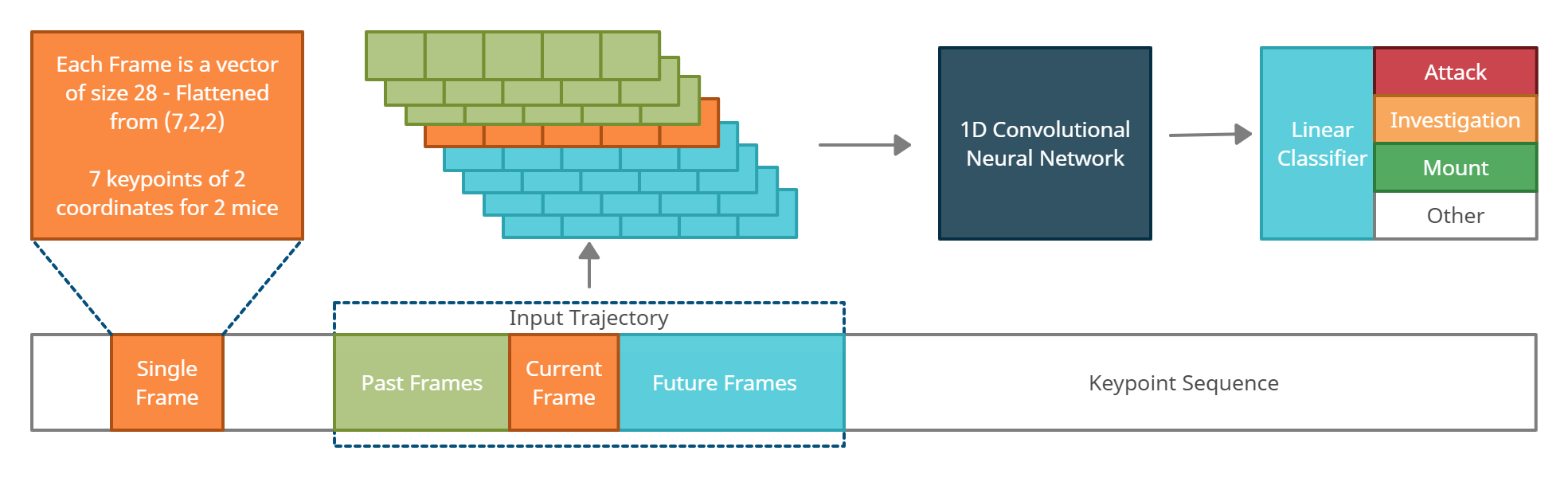}
    \caption{\textbf{Sequence Classification Setup.} Sequence information from past, present, and future frames may be used to predict the observed behavior label on the current frame. Here, we show a 1D convolutional neural network, but in general any model may be used.}
    \label{fig:neuralnetflow}
\end{figure*}

\begin{figure*}
    \centering
    \includegraphics[width=\linewidth]{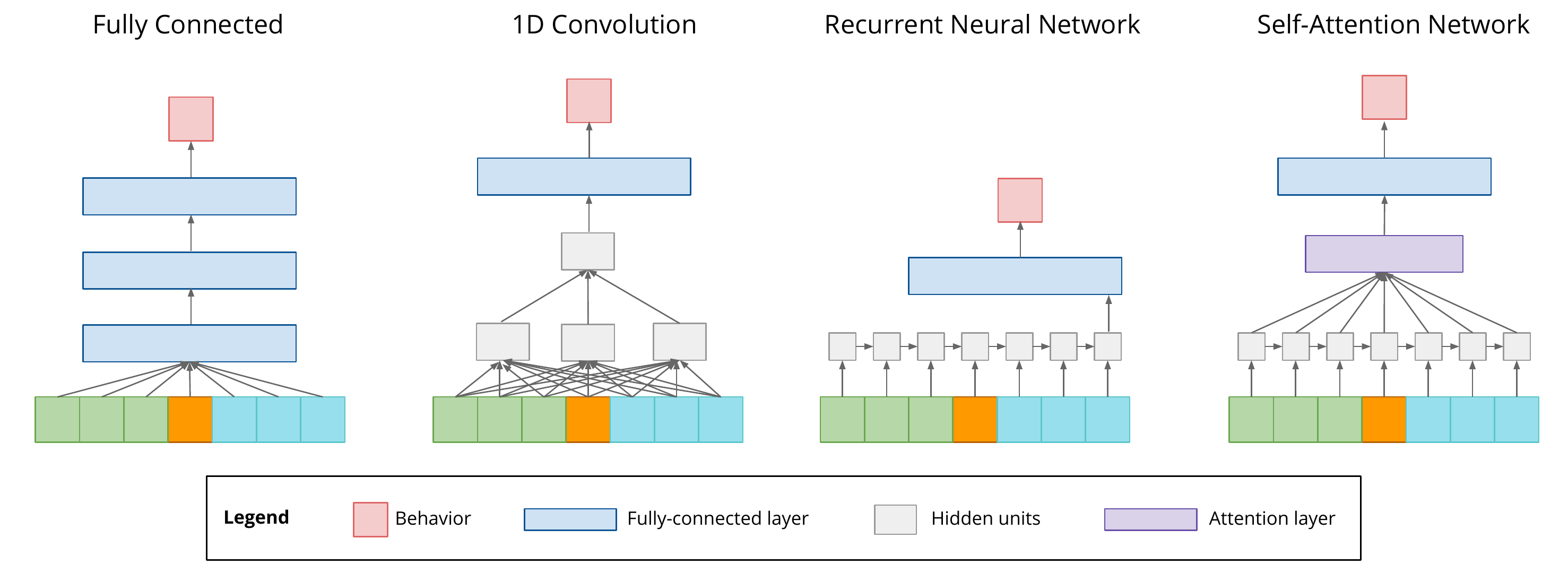}
    \caption{\textbf{Baseline models.} Different baseline setups we evaluated for behavior classification. The input frame coloring follows the same convention as Figure~\ref{fig:neuralnetflow}: past frames in green, current frame in orange, and future frames in cyan.}
    \label{fig:model_setup}
\end{figure*}

\subsection{Baseline Model Architectures}~\label{sec:baseline_architectures}
Our goal is to estimate the behavior labels in each frame from trajectory data, and we use information from both past and future frames for this task (Figure~\ref{fig:neuralnetflow}). To establish baseline classification performance on CalMS21, we explored a family of neural network-based classification approaches (Figure~\ref{fig:model_setup}). All models were trained using categorical cross entropy loss~\cite{goodfellow2016machine} on Task 1: Classic Classification, using an 80/20 split of the train split into training and validation sets during development. We report results on the full training set after fixing model hyperparameters.

Among the explored architectures, we obtained the highest performance using the 1D convolutional neural network (Table \ref{table:task1architectures}). We therefore used this architecture for baseline evaluations in all subsequent sections.

Hyperparameters we considered includes the number of input frames, the numer of frame skips, the number of units per layer, and the learning rate. Settings of these parameters may be found in the project code and the appendix. The baseline with task programming models use the same hyperparameters as the baseline. The task programming model is trained on the unlabeled set only, and the learned features are concatenated with the keypoints at each frame. 

\begin{wraptable}{r}{7.0cm}
\vspace{-0.5in}
    \centering
    \scalebox{0.95}{
    \begin{tabular}{l|c|c}
        \toprule[0.2em]
        Method  & Average F1 & MAP \\        
        \toprule[0.2em]
        Fully Connected & $.659 \pm .005$ & $.726 \pm .004$ \\
        LSTM & $.675 \pm .011$ & $.712 \pm .013$\\
        Self-Attention & $.610 \pm .028$ & $.644 \pm .018$\\ 
        1D Conv Net & $.793 \pm .011$ & $.856 \pm .010$\\
        \bottomrule[0.1em]
    \end{tabular}
    }
    \caption{Class-averaged results on Task 1 (attack, investigation, mount) for different baseline model architectures. The value is shown of the mean and standard deviation over 5 runs. }
    \vspace{-0.2in}
    \label{table:task1architectures}   
\end{wraptable}

\subsection{Task 1 Classic Classification Results}~\label{sec:task1_results}
\begin{figure*}[t]
        \centering
    \includegraphics[width=\linewidth]{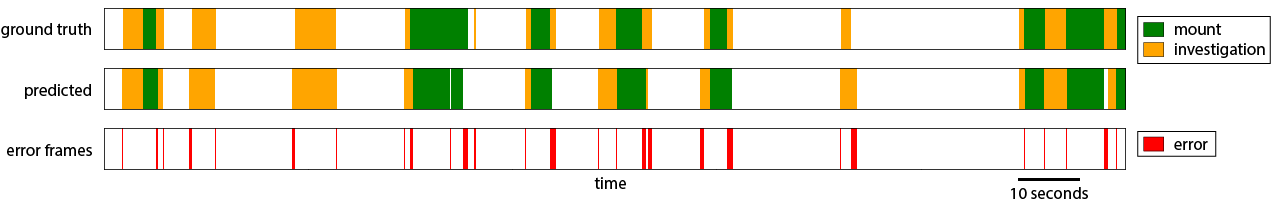}
    \caption{Example of errors from a sequence of behaviors from Task 1.}
    \label{fig:task1error}
    \vspace{-0.1in}
\end{figure*}

\textbf{Baseline Models.} We used the 1D convolutional neural network model outlined above (Figure \ref{fig:neuralnetflow}) to predict attack, investigation, and mount behaviors in two settings: using raw pose trajectories, and using trajectories plus features learned from the unlabeled set using task programming (Table \ref{table:task1metrics}). We found that including task programming features improved model performance. Many prediction errors of the baseline models were localized around behavior transition boundaries (Figure \ref{fig:task1error}). These errors may arise in part from annotator noise in the human generated labels of behaviors. An analysis of such intra- (and inter-) annotator variability is found in~\cite{segalin2020mouse}.

\textbf{Task1 Top-1 Entry.} We also include the top-1 entry in Task 1 of our dataset at MABe 2021 as part of our benchmark (Table \ref{table:task1metrics}). This model starts from an egocentric representation of the data; in a preprocessing stage, features are computed based on distances, velocities, and angles between coordinates relative to the agents' orientations. Furthermore, a PCA embedding of pairwise distances of all coordinates of both individuals is given as input to the model.

The model architecture is based on~\cite{oord2018representation} with three main components. First, the embedder network consists of several residual blocks~\cite{he2016deep} of non-causal 1D convolutional layers. Next, the contexter network is a stack of residual blocks with causal 1D convolutional layers. Finally, a fully connected residual block with multiple linear classification heads computes the class probabilities for each behavior.  Additional inputs such as a learned embedding for the annotator (see Section~\ref{sec:task2_results}) and absolute spatial and temporal information are directly fed into this final component.

The Task 1 top-1 model was trained in a semi-supervised fashion, using the normalized temperature-scaled cross-entropy loss~\cite{oord2018representation,chen2020simple} for all samples and the categorical cross-entropy loss for labeled samples. During training, sequences were sampled from the data proportional to their length with a 3:1 split of unlabeled/labeled sequences. Linear transformations that project the contexter component's outputs into the future are learned jointly with the model, as described in~\cite{oord2018representation}.  This unsupervised loss component regularizes the model by encouraging the model to learn a representation that is predictive of future behavior. A single model was trained jointly for all three tasks of the challenge, with all parameters being shared among the tasks, except for the final linear classification layers. The validation loss was monitored during training, and a copy of the parameters with the lowest validation loss was stored for each task individually.

\begin{table}[t]
    \centering
    \scalebox{0.95}{
    \begin{tabular}{l|ccc|c|c}
        \toprule[0.2em]
        \multirow{3}{*}{Method} & \multicolumn{3}{c|}{Data Used During Training} & \multirow{3}{*}{Average F1} & \multirow{3}{*}{MAP} \\
        & Task 1 & Unlabeled Set & All Tasks & & \\
        & (train split) &  & (all splits) & & \\        
        \toprule[0.2em]
        Baseline & \checkmark & &  & $.793 \pm .011$ & $.856 \pm .010$ \\
        Baseline w/ task prog & \checkmark & \checkmark &  & $.829 \pm .004$ & $.889 \pm .004$\\
        MABe 2021 Task 1 Top-1 & \checkmark & \checkmark & \checkmark & $.864 \pm .011$ & $.914 \pm .009$ \\
        \bottomrule[0.1em]
    \end{tabular}
    }
    \caption{Class-averaged results on Task 1 (attack, investigation, mount; mean $\pm$ standard deviation over 5 runs.) See appendix for per class results. The ``All Tasks" column indicates that the model was jointly trained on all three tasks, and ``all splits" indicates that both labeled train set and trajectory from unlabeled test set are used. }\label{table:task1metrics}   
    \vspace{-0.1in}
\end{table}

\begin{table}[t]
    \centering
    \scalebox{0.95}{
    \begin{tabular}{l|ccc|c|c}
        \toprule[0.2em]
        \multirow{3}{*}{Method} & \multicolumn{3}{c|}{Data Used During Training} & \multirow{3}{*}{Average F1} & \multirow{3}{*}{MAP} \\
        & Task 1 \& 2 & Unlabeled Set & All Tasks & & \\
        & (train split) &  &  (all splits) & & \\        
        \toprule[0.2em]
        Baseline & \checkmark & &  & $.754 \pm .005$ & $.813 \pm .003$  \\
        Baseline w/ task prog & \checkmark & \checkmark &  & $.774 \pm .006$ & $.835 \pm .005$ \\
        MABe 2021 Task 2 Top-1 & \checkmark & \checkmark & \checkmark & $.809 \pm .015$ & $.857 \pm .007$ \\
        \bottomrule[0.1em]
    \end{tabular}
    }
    \caption{Class-averaged and annotator-averaged results on Task 2 (attack, investigation, mount; mean $\pm$ standard deviation over 5 runs). The ``All Tasks" column indicates that the model was jointly trained on all three tasks, and ``all splits" indicates that both labeled train set and trajectory from unlabeled test set are used. See appendix for per class and per annotator results. }\label{table:task2metrics}   
    \vspace{-0.1in}
\end{table}

\subsection{Task 2 Annotation Style Transfer Results} ~\label{sec:task2_results}
\textbf{Baseline Model.} Similar to Task 1, Task 2 involves classifying attack, investigation, and mount frames. However, in this task, our goal is to capture the particular annotation style of different individuals. This step is important in identifying sources of discrepancy in behavior definitions between datasets or labs. Given the limited training set size in Task 2 (only 6 videos for each annotator), we used the model trained on Task 1 as a pre-trained model for the baseline experiments in Task 2, to leverage the larger training set from Task 1. The performances are in Table~\ref{table:task2metrics}, with per-annotator results in the appendix.

\textbf{Task2 Top-1 Entry.} The top MABe submission for Task 2 re-used the model architecture and training schema from Task 1, described in Section~\ref{sec:task1_results}. To address different annotation styles in Task 2, a learned annotator embedding was concatenated to the outputs of the contexter network. This embedding was initialized as a diagonal matrix such that initially, each annotator is represented by a one-hot vector. This annotator matrix is learnable, so the network can learn to represent similarities in the annotation styles. Annotators 3 and 4 were found to be similar to each other, and different from annotators 1, 2, and 5. The learned annotator matrix is provided in the appendix.

\begin{table}[t]
    \centering
    \scalebox{0.95}{
    \begin{tabular}{l|ccc|c|c}
        \toprule[0.2em]
        \multirow{3}{*}{Method} & \multicolumn{3}{c|}{Data Used During Training} & \multirow{3}{*}{Average F1} & \multirow{3}{*}{MAP} \\
        & Task 1 & Task 3  & Unlabeled Set & & \\
        & (train split) & (train split) & & \\
        \toprule[0.2em]
        Baseline & \checkmark & \checkmark &  & $0.338 \pm .004$ & $.317 \pm .005$ \\
        Baseline w/ task prog & \checkmark & \checkmark & \checkmark & $.328 \pm .009$ & $.320 \pm .009$ \\
        \multirow{2}{*}{MABe 2021 Task 3 Top-1} &  & \multirow{2}{*}{\checkmark} &   & $.319 \pm .025$ & \multirow{2}{*}{$.352 \pm .023$}\\
        & & & & ($.363 \pm .020$) \\
        \bottomrule[0.1em]
    \end{tabular}
    }
    \caption{Class-averaged results on Task 3 over the 7 behaviors of interest (mean $\pm$ standard deviation over 5 runs.) The average F1 score in brackets corresponds to improvements with threshold tuning. See appendix for per class results.}
    \label{table:task3metrics}   
    \vspace{-0.1in}
\end{table}

\subsection{Task 3 New Behaviors Results}~\label{sec:task3_results}
\textbf{Baseline Model.} Task 3 is a set of data-limited binary classification problems with previously unseen behaviors. Although these behaviors do occur in the Task 1 and Task 2 datasets, they are not labeled. The challenges in this task arise from both the low amount of training data for each new behavior and the high class imbalance, as seen in Figure ~\ref{fig:data_description}. 

For this task, we used our trained Task 1 baseline model as a starting point. Due to the small size of the training set, we found that models that did not account for class imbalance performed poorly. We therefore addressed class imbalance in our baseline model by replacing our original loss function with a weighted cross-entropy loss in which we scaled the weight of the under-represented class by the number of training frames for that class. Results for Task 3 are provided in Table \ref{table:task3metrics}. We found classifier performance to depend both on the percentage of frames during which a behavior was observed, and on the average duration of a behavior bout, with shorter bouts having lower classifier performance. 

\textbf{Task3 Top-1 Entry.} The model for the top entry in Task 3 of the MABe Challenge was inspired by spatial-temporal graphs that have been used for skeleton-based action recognition algorithms in human pose datasets. In particular, MS-G3D~\cite{liu2020disentangling} is an effective algorithm for extracting multi-scale spatial-temporal features and long-range dependencies. The MS-G3D model is composed of a stack of multiple spatial-temporal graph convolution blocks, followed by a global average pooling layer and a softmax classifier~\cite{liu2020disentangling}.

The spatial graph for Task 3 is constructed using the detected pose keypoints, with a connection added between the necks of the two mice. The inputs are normalized following ~\cite{yan2018spatial}. MS-G3D is then trained in a fully supervised fashion on the train split of Task 3. Additionally, the model is trained with data augmentation based on rotation.


\section{Discussion}~\label{sec:discussion}
We introduce CalMS21, a new dataset for detecting the actions of freely behaving mice engaged in naturalistic social interactions in a laboratory setting. The released data include over 70 hours of tracked poses from pairs of mice, and over 10 hours of manual, frame-by-frame annotation of animals' actions. 
Our dataset provides a new way to benchmark the performance of multi-agent behavior classifiers. In addition to reducing human effort, automated behavior classification can lead to more objective, precise, and scalable measurements compared to manual annotation \cite{anderson2014toward, dell2014automated}. Furthermore, techniques studied on our dataset can be potentially applied to other multi-agent datasets, such as those for sports analytics and autonomous vehicles. 

In addition to the overall goal of supervised behavior classification, we emphasize two specific problems where we see a need for further investigation. The first of these is the utility of behavior classifiers for comparison of annotation style between different individuals or labs, most closely relating to our Task 2 on annotation style transfer. The ability to identify sources of inter-annotator disagreement is important for the reproducibility of behavioral results, and we hope that this dataset will foster further investigation into the variability of human-defined behavior annotations. A second problem of interest is the automated detection of new behaviors of interest given limited training data. This is especially important for the field of automated behavior analysis, as few-shot training of behavior classifiers would enable researchers to use supervised behavior classification as a tool to rapidly explore and curate large datasets of behavioral videos. 

Alongside manually annotated trajectories provided for classifier training and testing, we include a large set of unlabeled trajectory data from 282 videos. The unlabeled dataset may be used to improve the performance of supervised classifiers, for example by learning self-supervised representations of trajectories~\cite{sun2020task}, or it may be used on its own for the development of unsupervised methods for behavior discovery or trajectory forecasting. We note that trajectory forecasting is a task that is of interest to other fields studying multi-agent behavior, including self-driving cars and sports analytics. We hope that our dataset can provide an additional domain with which to test these models. In addition, unsupervised behavior analysis may be capable of identifying a greater number of behaviors than a human annotator would be able to annotate reliably. Recent single-animal work has shown that unsupervised pose analyses can enable the detection of subtle differences between experimental conditions~\cite{wiltschko2020revealing}. A common problem in unsupervised analysis is evaluating the quality of the learned representation. Therefore, an important topic to be addressed in future work is the development of appropriate challenge tasks to evaluate the quality of unsupervised representations of animal movements and actions, beyond comparison with human-defined behaviors.

\textbf{Broader Impact.} In recent years, animal behavior analysis has emerged as a powerful tool in the fields of biology and neuroscience, enabling high-throughput behavioral screening in hundreds of hours of behavioral video~\cite{branson2009high}. Prior to the emergence of these tools, behavior analysis relied on manual frame-by-frame annotation of animals' actions, a process which is costly, subjective, and arduous for domain experts. The increased throughput enabled by automation of behavior analysis has seen applications in neural circuit mapping~\cite{robie2017mapping, cande2018optogenetic}, computational drug development~\cite{wiltschko2020revealing}, evolution~\cite{hernandez2020framework}, ecology~\cite{dell2014automated}, and studies of diseases and disorders~\cite{machado2015quantitative,hong2015automated,wiltschko2020revealing}. In releasing this dataset, our hope is to establish community benchmarks and metrics for the evaluation of new computational behavior analysis tools, particularly for social behaviors, which are particularly challenging to investigate due to their heterogeneity and complexity.

In addition to behavioral neuroscience, behavior modeling is of interest to diverse fields, including autonomous vehicles, healthcare, and video games. While behavior modeling can help accelerate scientific experiments and lead to useful applications, some applications of these models to human datasets, such as for profiling users or for conducting surveillance, may require more careful consideration. Ultimately, users of behavior models need to be aware of potentially negative societal impacts caused by their application. 

\textbf{Future Directions.} In this dataset release, we have opted to emphasize behavior classification from keypoint-based animal pose estimates. However, it is possible that video data could further improve classifier performance. Since we have also released accompanying video data to a subset of CalMS21, an interesting future direction would be to determine the circumstances under which video data can improve behavior classification. Additionally, our dataset currently focuses on top-view tracked poses from a pair of interacting mice. For future iterations, including additional organisms, experimental settings, and task designs could further help benchmark the performance of behavior classification models. Finally, we value any input from the community on CalMS21 and you can reach us at \href{mailto:mabe.workshop@gmail.com}{mabe.workshop@gmail.com}.

\section{Acknowledgements}

We would like to thank the researchers at the David Anderson Research Group at Caltech for this collaboration and the recording and annotation of the mouse behavior datasets, in particular, Tomomi Karigo, Mikaya Kim, Jung-sook Chang, Xiaolin Da, and Robert Robertson. We are grateful to the team at AICrowd for the support and hosting of our dataset challenge, as well as Northwestern University and Amazon Sagemaker for funding our challenge prizes. This work was generously supported by the Simons Collaboration on the Global Brain grant 543025 (to PP), NIH Award \#K99MH117264 (to AK), NSF Award \#1918839 (to YY), NSERC Award \#PGSD3-532647-2019 (to JJS), as well as a gift from Charles and Lily Trimble (to PP).

\newpage

\begin{small}
\bibliographystyle{plain}
\bibliography{references}
\end{small}

\newpage

\newpage


\appendix

\section*{Appendix for CalMS21}

The sections of our appendix are organized as follows:
\begin{itemize}
    \item[\textbullet] \hyperref[sec:license]{\textbf{\textcolor{blue}{Section A}}} contains dataset hosting and licensing information.
    \item[\textbullet] \hyperref[sec:datasheets]{\textbf{\textcolor{blue}{Section B}}} contains dataset documentation and intended uses for CalMS21, following the format of the Datasheet for Datasets\cite{gebru2018datasheets}.
    \item[\textbullet] \hyperref[sec:dataformat]{\textbf{\textcolor{blue}{Section C}}} describes the data format (.json).    
    \item[\textbullet] \hyperref[sec:datasetpreparation]{\textbf{\textcolor{blue}{Section D}}} describes how animal behavior data was recorded and processed.
    \item[\textbullet] \hyperref[sec:evaluation]{\textbf{\textcolor{blue}{Section E}}} shows the evaluation metrics for CalMS21, namely the F1 score and Average Precision.
    \item[\textbullet] \hyperref[sec:implementationdetails]{\textbf{\textcolor{blue}{Section F}}} contains additional implementation details of our models.
    \item[\textbullet] \hyperref[sec:additionalresults]{\textbf{\textcolor{blue}{Section G}}} provides additional evaluation results.
    \item[\textbullet] \hyperref[sec:benchmarkreproducibility]{\textbf{\textcolor{blue}{Section H}}} addresses benchmark model reproducibility, following the format of the ML Reproducibility Checklist\cite{pineau2020improving}.    
\end{itemize}


\section{CalMS21 Hosting and Licensing}~\label{sec:license}

The CalMS21 dataset is available at \href{https://sites.google.com/view/computational-behavior/our-datasets/calms21-dataset}{https://sites.google.com/view/computational-behavior/our-datasets/calms21-dataset} (DOI: \href{https://doi.org/10.22002/D1.1991}{https://doi.org/10.22002/D1.1991}), and is distributed under a CreativeCommons Attribution/Non-Commercial/Share-Alike license (CC-BY-NC-SA).

CalMS21 is hosted via the Caltech Research Data Repository at data.caltech.edu. This is a static dataset, meaning that any changes (such as new tasks, new experimental data, or improvements to pose estimates) will be released as a new entity; these updates will typically accompany new iterations of the MABe Challenge. News of any such updates will be posted both to the dataset website \href{https://sites.google.com/view/computational-behavior/our-datasets/calms21-dataset}{https://sites.google.com/view/computational-behavior/our-datasets/calms21-dataset} and on the data repository page at \href{https://data.caltech.edu/records/1991}{https://data.caltech.edu/records/1991}. 

Code for all baseline models is available at \href{https://gitlab.aicrowd.com/aicrowd/research/mab-e/mab-e-baselines}{https://gitlab.aicrowd.com/aicrowd/research/mab-e/mab-e-baselines}, and is distributed under the MIT License. We as authors bear all responsibility in case of violation of rights.

Code for the top entry for Tasks 1 \& 2 will be released by BW under the MIT License. The code for Task 3 will not be made publicly available.

\section{CalMS21 Documentation and Intended Uses}~\label{sec:datasheets}
This section follows the format of the Datasheet for Datasets\cite{gebru2018datasheets}.
\dssectionheader{Motivation}

\dsquestionex{For what purpose was the dataset created?}{Was there a specific task in mind? Was there a specific gap that needed to be filled? Please provide a description.}

\dsanswer{Automated animal pose estimation has become an increasingly popular tool in the neuroscience community, fueled by the publication of several easy-to-train animal pose estimation systems. Building on these pose estimation tools, pose-based approaches to supervised or unsupervised analysis of animal behavior are currently an area of active research. New computational approaches for automated behavior analysis are probing the detailed temporal structure of animal behavior, its relationship to the brain, and how both brain and behavior are altered in conditions such as Parkinson’s, PTSD, Alzheimer’s, and autism spectrum disorders. Due to a lack of publicly available animal behavior datasets, most new behavior analysis tools are evaluated on their own in-house data. There are no established community standards by which behavior analysis tools are evaluated, and it is unclear how well available software can be expected to perform in new conditions, particularly in cases where training data is limited. Labs looking to incorporate these tools in their experimental pipelines therefore often struggle to evaluate available analysis options, and can waste significant effort training and testing multiple systems without knowing what results to expect.

The Caltech Mouse Social 2021 (CalMS21) dataset is a new animal tracking, pose, and behavioral dataset, intended to a) serve as a benchmark dataset for comparison of behavior analysis tools, and establish community standards for evaluation of behavior classifier performance b) highlight critical challenges in computational behavior analysis, particularly pertaining to leveraging large, unlabeled datasets to improve performance on supervised classification tasks with limited training data, and c) foster interaction between behavioral neuroscientists and the greater machine learning community.
}

\dsquestion{Who created this dataset (e.g., which team, research group) and on behalf of which entity (e.g., company, institution, organization)?}

\begin{figure*}
    \centering
  \includegraphics[width=\linewidth]{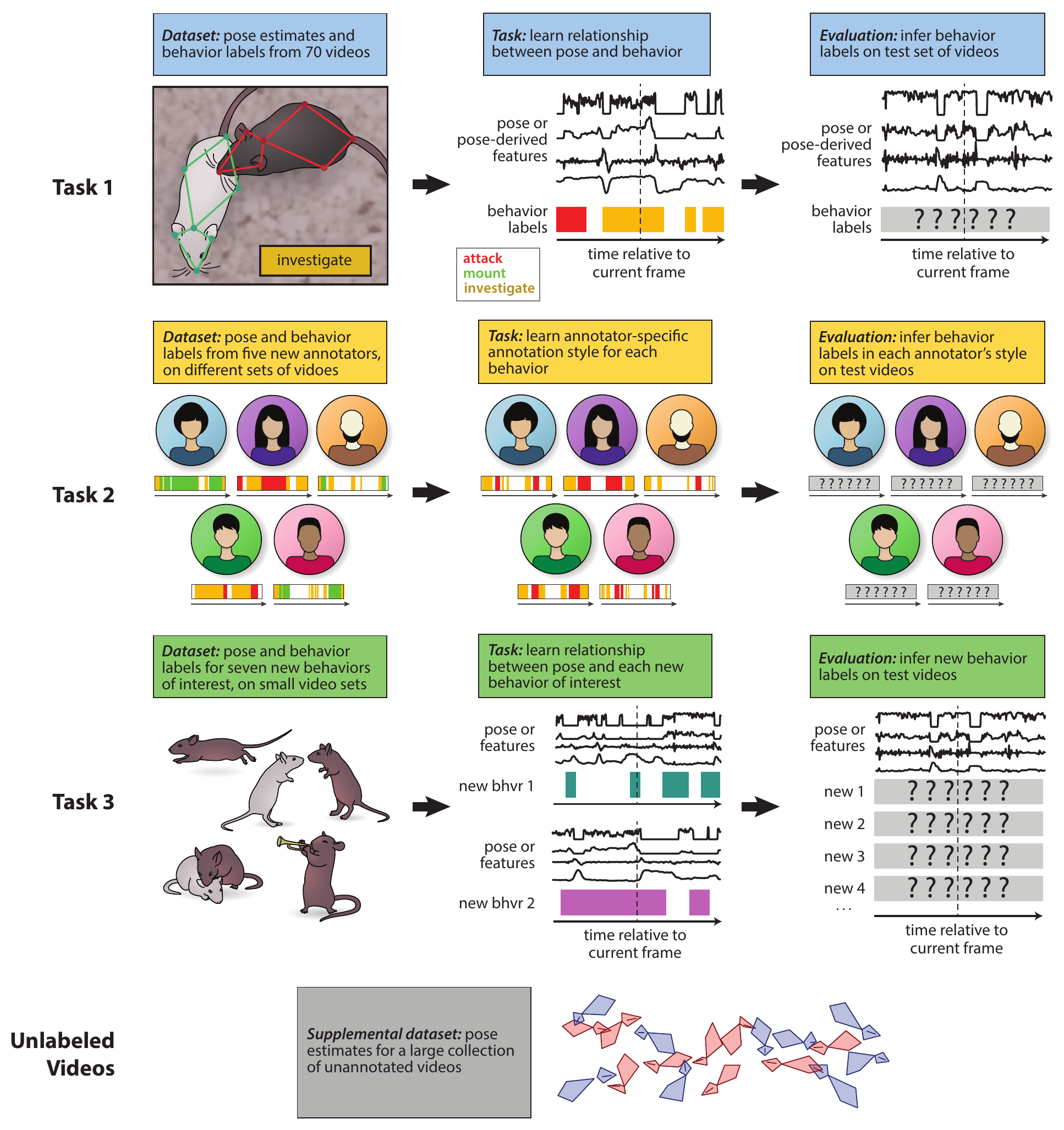}
  \caption{{\bf Summary of Tasks.} Visual summary of datasets, tasks, and evaluations for the three tasks defined in CalMS21.}
  \label{fig:task_summary}
\end{figure*}

\dsanswer{The CalMS21 dataset was created as a collaborative effort between the laboratories of David J Anderson, Yisong Yue, and Pietro Perona at Caltech, and Ann Kennedy at Northwestern. Videos of interacting mice were produced and manually annotated by Tomomi Karigo and other members of the Anderson lab. The video dataset was tracked, curated and screened for tracking quality by Ann Kennedy and Jennifer J. Sun, with pose estimation performed using version 1.7 of the Mouse Action Recognition System (MARS). The dataset tasks (Figure~\ref{fig:task_summary}) were designed by Ann Kennedy and Jennifer J. Sun, with input from Pietro Perona and Yisong Yue.
}

\dsquestionex{Who funded the creation of the dataset?}{If there is an associated grant, please provide the name of the grantor and the grant name and number.}

\dsanswer{Acquisition of behavioral data was supported by NIH grants R01 MH085082 and R01 MH070053, Simons Collaboration on the Global Brain Foundation award no. 543025 (to DJA and PP), as well as a HFSP Long-Term Fellowship (to TK). Tracking, curation of videos, and task design was funded by NIMH award \#R00MH117264 (to AK), NSF Award \#1918839 (to YY), and NSERC Award \#PGSD3-532647-2019 (to JJS).
}

\dsquestion{Any other comments?}

\dsanswer{None.}

\bigskip
\dssectionheader{Composition}

\dsquestionex{What do the instances that comprise the dataset represent (e.g., documents, photos, people, countries)?}{ Are there multiple types of instances (e.g., movies, users, and ratings; people and interactions between them; nodes and edges)? Please provide a description.}

\dsanswer{The core element of this dataset, called a \textit{sequence}, captures the tracked postures and actions of two mice interacting in a standard resident-intruder assay filmed from above at 30Hz and manually annotated on a frame-by-frame basis for one or more behaviors. The resident in these assays is always a male mouse from strain C57Bl/6J, or from a transgenic line with C57Bl/6J background. The intruder is a male or female BALB/c mouse. Resident mice may be either group-housed or single-housed, and either socially/sexually naive or experienced (all factors that impact the types of social behaviors animals show in this assay.)

The core element of a \textit{sequence} is called a \textit{frame}; this refers to the posture of both animals on a particular frame of video, as well as one or more labels indicating the type of behavior being performed on that frame (if any).

The dataset is divided into four sub-sets: three collections of sequences associated with Tasks 1, 2, and 3 of the MABe Challenge, and a fourth "Unlabeled" collection of sequences that have only the \textit{keypoint} elements with no accompanying \textit{annotations} or \textit{annotator-id} (see "What data does each instance consist of?" for explanation of these values.) Tasks 1-3 are split into train and test sets. Tasks 2 and 3 are also split by \textit{annotator-id} (Task 2) or behavior (Task 3).
}

\dsquestion{How many instances are there in total (of each type, if appropriate)?}

\dsanswer{
Instances for each dataset are shown in table ~\ref{table:framecounts}, divided into train and test sets. Number of instances is reported as both \textit{frames} and \textit{sequences}, where frames within a sequence are temporally contiguous and sampled at 30Hz (and hence not true statistically independent observations).

\begin{table}
    \begin{center}
         \begin{tabular}{| c | l | rc | rc |} 
         \hline
         \multirow{2}{*}{Task} & \multirow{2}{*}{Category} & \multicolumn{2}{c|}{Training set} & \multicolumn{2}{c|}{Test set} \\
        & & Frames & Sequences & Frames & Sequences \\
         \hline
         \hline
        Task 1 & -- & 507,738 & 70 & 262,107 & 19  \\
        \hline
        \hline
        \multirow{5}{*}{Task 2} & Annotator 1 & 139,112 & 6 & 286,810 & 13  \\
               & Annotator 2 & 135,623 & 6 & 150,919 & 6  \\
               & Annotator 3 & 107,420 & 6 &  77,079 & 4  \\
               & Annotator 4 & 108,325 & 6 &  76,174 & 4  \\
               & Annotator 5 &  92,383 & 6 & 364,007 & 20  \\
        \hline
        \hline
        \multirow{7}{*}{Task 3} & Approach & 20,624 & 3 & 126,468 & 25  \\
               & Disengaged    & 35,751 & 2 &  19,088 & 1  \\
               & Grooming      & 45,174 & 2 & 156,664 & 13  \\
               & Intromission  & 19,200 & 1 &  55,218 & 10  \\
               & Mount-attempt & 46,847 & 4 &  85,836 & 12  \\
               & Sniff-face    & 19,244 & 3 & 251,793 & 47  \\
               & White rearing & 36,181 & 2 &  17,939 & 1  \\
        \hline
        \end{tabular}
    \end{center}
    \caption{Training and test set instances counts for each task and category.}
\label{table:framecounts}
\end{table}

}

\dsquestionex{Does the dataset contain all possible instances or is it a sample (not necessarily random) of instances from a larger set?}{ If the dataset is a sample, then what is the larger set? Is the sample representative of the larger set (e.g., geographic coverage)? If so, please describe how this representativeness was validated/verified. If it is not representative of the larger set, please describe why not (e.g., to cover a more diverse range of instances, because instances were withheld or unavailable).}

\dsanswer{The assembled dataset presented here was manually curated from a large, unreleased repository of mouse behavior videos collected across several years by multiple members of the Anderson lab. Only videos of naturally occurring (not optogenetically or chemogenetically evoked) behavior were included. Selection criteria are described in the "Collection Process" section.

As a result of our selection criteria, the videos included in the Tasks 1-3 datasets may not be fully representative of mouse behavior in the resident-intruder assay: videos with minimal social interactions (when the resident ignored or avoided the intruder) were omitted in favor of including a greater number of examples of the annotated behaviors of interest.
}

\dsquestionex{What data does each instance consist of? “Raw” data (e.g., unprocessed text or images) or features?}{In either case, please provide a description.}

\dsanswer{Each sequence has three elements. 1) \textit{Keypoints} are the locations of seven body parts (the nose, left and right ears, base of neck, left and right hips, and base of tail) on each of two interacting mice. Keypoints are estimated using the Mouse Action Recognition System (MARS). 2) \textit{Annotations} are manual, frame-wise labels of an animal's actions, for example attack, mounting, and close investigation. Depending on the behaviors annotated, only between a few percent and up to half of frames will have an annotated action; frames that do not have an annotated action are labeled as \textit{other}. The \textit{other} label should not be taken to indicate that no behaviors are happening, and it should not be considered a true label category for purposes of classifier performance evaluation. 3) \textit{Annotator-id} is a unique numeric ID indicating which (anonymized) human annotator produced the labels in \textit{Annotations}. This ID is provided primarily for use in Task 2 of the MABe Challenge, which pertains to annotator style capture.

Note that this dataset does not include the original raw videos from which pose estimates were produced. This is because the objective of releasing this dataset was to determine the accuracy with which animal behavior could be detected using tracked keypoints alone.
}

\dsquestionex{Is there a label or target associated with each instance?}{If so, please provide a description.}

\dsanswer{In the Task 1, Task 2, and Task 3 datasets, the \textit{annotation} field for a given behavior sequence consists of frame-wise labels of animal behaviors. Note that only a minority of frames have behavior labels; remaining frames are labeled as \textit{other}. Only a small number of behaviors were tracked by human annotators (most typically \textit{attack}, \textit{mount}, and \textit{close investigation}), therefore frames labeled as \textit{other} are not a homogeneous category, but may contain diverse other behaviors.

The "Unlabeled" collection of sequences has no labels, and instead contains only keypoint tracking data.
}

\dsquestionex{Is any information missing from individual instances?}{If so, please provide a description, explaining why this information is missing (e.g., because it was unavailable). This does not include intentionally removed information, but might include, e.g., redacted text.}

\dsanswer{There is no missing data (beyond what was intentionally omitted, eg in the Unlabeled category.)
}

\dsquestionex{Are relationships between individual instances made explicit (e.g., users’ movie ratings, social network links)?}{If so, please describe how these relationships are made explicit.}

\dsanswer{Each instance (\textit{sequence}) is to be treated as an independent observation with no relationship to other instances in the dataset. In almost all cases, the identities of the interacting animals are unique to each sequence, and this information is not tracked in the dataset.
}

\dsquestionex{Are there recommended data splits (e.g., training, development/validation, testing)?}{If so, please provide a description of these splits, explaining the rationale behind them.}

\dsanswer{The dataset includes a recommended train/test split for Tasks 1, 2, and 3. In Tasks 2 and 3, the split was designed to provide a roughly consistent, small amount of training data for each sub-task. In Task 1, the split was manually selected so that the test set included sequences from a range of experimental conditions and dates.
}

\dsquestionex{Are there any errors, sources of noise, or redundancies in the dataset?}{If so, please provide a description.}

\dsanswer{Pose keypoints in this dataset are produced using automated pose estimation software (the Mouse Action Recognition System, MARS). While the entire dataset was manually screened to remove sequences with poor pose estimation, some errors in pose estimation and noise in keypoint placement still occur. These are most common on frames when the two animals are in close contact or moving very quickly.

In addition, manual annotations of animal behavior are inherently subjective, and individual annotators show some variability in the precise frame-by-frame labeling of behavior sequences. An investigation of within- and between-annotator variability is included in the MARS pre-print.
}

\dsquestionex{Is the dataset self-contained, or does it link to or otherwise rely on external resources (e.g., websites, tweets, other datasets)?}{If it links to or relies on external resources, a) are there guarantees that they will exist, and remain constant, over time; b) are there official archival versions of the complete dataset (i.e., including the external resources as they existed at the time the dataset was created); c) are there any restrictions (e.g., licenses, fees) associated with any of the external resources that might apply to a future user? Please provide descriptions of all external resources and any restrictions associated with them, as well as links or other access points, as appropriate.}

\dsanswer{The dataset is self-contained.
}

\dsquestionex{Does the dataset contain data that might be considered confidential (e.g., data that is protected by legal privilege or by doctor-patient confidentiality, data that includes the content of individuals non-public communications)?}{If so, please provide a description.}

\dsanswer{No.
}

\dsquestionex{Does the dataset contain data that, if viewed directly, might be offensive, insulting, threatening, or might otherwise cause anxiety?}{If so, please describe why.}

\dsanswer{No such material; dataset contains only tracked posture keypoints (no video or images) and text labels pertaining to mouse social behaviors.
}

\dsquestionex{Does the dataset relate to people?}{If not, you may skip the remaining questions in this section.}

\dsanswer{No.
}

\dsquestionex{Does the dataset identify any subpopulations (e.g., by age, gender)?}{If so, please describe how these subpopulations are identified and provide a description of their respective distributions within the dataset.}

\dsanswer{n/a
}

\dsquestionex{Is it possible to identify individuals (i.e., one or more natural persons), either directly or indirectly (i.e., in combination with other data) from the dataset?}{If so, please describe how.}

\dsanswer{n/a
}

\dsquestionex{Does the dataset contain data that might be considered sensitive in any way (e.g., data that reveals racial or ethnic origins, sexual orientations, religious beliefs, political opinions or union memberships, or locations; financial or health data; biometric or genetic data; forms of government identification, such as social security numbers; criminal history)?}{If so, please provide a description.}

\dsanswer{n/a
}

\dsquestion{Any other comments?}

\dsanswer{
A subset of videos in Task 1 and the Unlabeled dataset are from animals that have been implanted with a head-mounted microendoscope or optical fiber (for fiber photometry.) Because the objective of this dataset is to learn to recognize behavior in a manner that is invariant to experimental setting, the precise preparation of the resident and intruder mice (including age, sex, past experiences, and presence of neural recording devices) is not provided in the dataset.
}

\bigskip
\dssectionheader{Collection Process}

\dsquestionex{How was the data associated with each instance acquired?}{Was the data directly observable (e.g., raw text, movie ratings), reported by subjects (e.g., survey responses), or indirectly inferred/derived from other data (e.g., part-of-speech tags, model-based guesses for age or language)? If data was reported by subjects or indirectly inferred/derived from other data, was the data validated/verified? If so, please describe how.}

\dsanswer{\textit{Sequences} in the dataset are derived from video of pairs of socially interacting mice engaged in a standard resident-intruder assay. In this assay, a black (C57Bl/6J) male "resident" mouse is filmed in its home cage, and a white (BALB/c) male or female "intruder" mouse is manually introduced to the cage by an experimenter. The animals are then allowed to freely interact for between 1-2 and 10 minutes. If there is excessive fighting (injury to either animal) the assay is stopped and that trial is discarded. Resident mice typically undergo several (3-6) resident-intruder assays per day with different intruder animals.

Poses of both mice were estimated from top-view video using MARS, and pose sequences were cropped to only include frames where both animals were present in the arena. Manual, frame-by-frame annotation of animals' actions were performed from top- and front-view video by trained experts.
}

\dsquestionex{What mechanisms or procedures were used to collect the data (e.g., hardware apparatus or sensor, manual human curation, software program, software API)?}{How were these mechanisms or procedures validated?}

\dsanswer{Video of the resident-intruder assay was captured at 30Hz using top- and front-view cameras (Point Grey Grasshopper3) recorded at 1024x570 (top) and 1280x500 (front) pixel resolution. Manual annotation was performed using custom software (either the Caltech Behavior Annotator (\href{https://github.com/pdollar/toolbox/blob/master/videos/behaviorAnnotator.m}{link}) or Bento (\href{https://github.com/neuroethology/bentoMAT}{link})) by trained human experts. All annotations were visually screened to ensure that the full sequence was annotated.
}

\dsquestion{If the dataset is a sample from a larger set, what was the sampling strategy (e.g., deterministic, probabilistic with specific sampling probabilities)?}

\dsanswer{The Task 1 dataset was chosen to match the training and test sets of behavior classifiers of MARS. These training and test sets, in turn, were sampled from among unpublished videos collected and annotated by a member of the Anderson lab. Selection criteria for inclusion were high annotation quality (as estimated by the individual who annotated the data) and annotation completeness; videos with diverse social behaviors (mounting and attack in addition to investigation) were favored. The Tasks 2 and 3 datasets were manually selected from among previously collected (unpublished) datasets, where selection criteria were for high annotation quality, annotation completeness, and sufficient number of behavior annotations. The Unlabeled dataset consists of videos from a subset of experiments in a recent publication\cite{karigo2021distinct}. The subset of experiments included in this dataset was chosen at random.
}

\dsquestion{Who was involved in the data collection process (e.g., students, crowdworkers, contractors) and how were they compensated (e.g., how much were crowdworkers paid)?}

\dsanswer{Behavioral data collection and annotation was performed by graduate student, postdoc, and technician members of the Anderson lab, as a part of other ongoing research projects in the lab. (No videos or annotations were explicitly generated for this dataset release.) Lab members are full-time employees of Caltech or HHMI, or are funded through independent graduate or postdoctoral fellowships, and their compensation was not dependent on their participation in this project.
}

\dsquestionex{Over what timeframe was the data collected? Does this timeframe match the creation timeframe of the data associated with the instances (e.g., recent crawl of old news articles)?}{If not, please describe the timeframe in which the data associated with the instances was created.}

\dsanswer{Data associated with this dataset was created and annotated between 2016 and 2020, with annotation typically occurring within a few weeks of creation. Pose estimation was performed later, with most videos processed in 2019-2020. This dataset was assembled from December 2020 - February 2021.
}

\dsquestionex{Were any ethical review processes conducted (e.g., by an institutional review board)?}{If so, please provide a description of these review processes, including the outcomes, as well as a link or other access point to any supporting documentation.}

\dsanswer{All experiments included here were performed in accordance with NIH guidelines and approved by the Institutional Animal Care and Use Committee (IACUC) and Institutional Biosafety Committee at Caltech. Review of experimental design by the IACUC follows the steps outlined in the NIH-published \href{https://grants.nih.gov/grants/olaw/Guide-for-the-Care-and-Use-of-Laboratory-Animals.pdf}{Guide for the Care and Use of Laboratory Animals}. All individuals performing behavioral experiments underwent animal safety training prior to data collection. Animals were maintained under close veterinary supervision, and resident-intruder assays were monitored in real time and immediately interrupted should either animal become injured during aggressive interactions.
}

\dsquestionex{Does the dataset relate to people?}{If not, you may skip the remaining questions in this section.}

\dsanswer{No.
}

\dsquestion{Did you collect the data from the individuals in question directly, or obtain it via third parties or other sources (e.g., websites)?}

\dsanswer{n/a
}

\dsquestionex{Were the individuals in question notified about the data collection?}{If so, please describe (or show with screenshots or other information) how notice was provided, and provide a link or other access point to, or otherwise reproduce, the exact language of the notification itself.}

\dsanswer{n/a
}

\dsquestionex{Did the individuals in question consent to the collection and use of their data?}{If so, please describe (or show with screenshots or other information) how consent was requested and provided, and provide a link or other access point to, or otherwise reproduce, the exact language to which the individuals consented.}

\dsanswer{n/a
}

\dsquestionex{If consent was obtained, were the consenting individuals provided with a mechanism to revoke their consent in the future or for certain uses?}{If so, please provide a description, as well as a link or other access point to the mechanism (if appropriate).}

\dsanswer{n/a
}

\dsquestionex{Has an analysis of the potential impact of the dataset and its use on data subjects (e.g., a data protection impact analysis) been conducted?}{If so, please provide a description of this analysis, including the outcomes, as well as a link or other access point to any supporting documentation.}

\dsanswer{n/a
}

\dsquestion{Any other comments?}

\dsanswer{None.
}

\bigskip
\dssectionheader{Preprocessing/cleaning/labeling}

\dsquestionex{Was any preprocessing/cleaning/labeling of the data done (e.g., discretization or bucketing, tokenization, part-of-speech tagging, SIFT feature extraction, removal of instances, processing of missing values)?}{If so, please provide a description. If not, you may skip the remainder of the questions in this section.}

\dsanswer{No preprocessing was performed on the \textit{sequence} data released in this dataset.
}

\dsquestionex{Was the “raw” data saved in addition to the preprocessed/cleaned/labeled data (e.g., to support unanticipated future uses)?}{If so, please provide a link or other access point to the “raw” data.}

\dsanswer{n/a
}

\dsquestionex{Is the software used to preprocess/clean/label the instances available?}{If so, please provide a link or other access point.}

\dsanswer{n/a
}

\dsquestion{Any other comments?}

\dsanswer{None.
}

\bigskip
\dssectionheader{Uses}

\dsquestionex{Has the dataset been used for any tasks already?}{If so, please provide a description.}

\dsanswer{Yes: this dataset was released to accompany the three tasks of the 2021 Multi-Agent Behavior (MABe) Challenge, posted \href{https://www.aicrowd.com/challenges/multi-agent-behavior-representation-modeling-measurement-and-applications}{here}. The challenge tasks are summarized as follows:
\begin{itemize}
    \item \textbf{Task 1, Classical Classification}: train supervised classifiers to detect instances of close investigation, mounting, and attack from labeled examples. All behaviors were annotated by the same individual.
    \item \textbf{Task 2, Annotation Style Transfer}: given limited training examples, train classifiers to reproduce the annotation style of five additional annotators for close investigation, mounting, and attack behaviors.
    \item \textbf{Task 3, Learning New Behavior}: given limited training examples, train classifiers to detect instances of seven additional behaviors (names of these behaviors were anonymized for this task.)
\end{itemize}
}

\dsquestionex{Is there a repository that links to any or all papers or systems that use the dataset?}{If so, please provide a link or other access point.}

\dsanswer{Papers that use or cite this dataset may be submitted by their authors for display on the CalMS21 website at \href{https://sites.google.com/view/computational-behavior/our-datasets/calms21-dataset}{https://sites.google.com/view/computational-behavior/our-datasets/calms21-dataset}
}

\dsquestion{What (other) tasks could the dataset be used for?}

\dsanswer{In addition to MABe Challenge Tasks 1-3, which can be studied with supervised learning, transfer learning, or few-shot learning techniques, the animal trajectories in this dataset could be used for unsupervised behavior analysis, representation learning, or imitation learning.
}

\dsquestionex{Is there anything about the composition of the dataset or the way it was collected and preprocessed/cleaned/labeled that might impact future uses?}{For example, is there anything that a future user might need to know to avoid uses that could result in unfair treatment of individuals or groups (e.g., stereotyping, quality of service issues) or other undesirable harms (e.g., financial harms, legal risks) If so, please provide a description. Is there anything a future user could do to mitigate these undesirable harms?}

\dsanswer{At time of writing there is no precise, numerical consensus definition of the mouse behaviors annotated in this dataset (and in fact even different individuals trained in the same research lab and following the same written descriptions of behavior can vary in how they define particular actions such as attack, as is evidenced in Task 2.) Future users should be aware of this limitation, and bear in mind that behavior annotations in this dataset may not always agree with the behavior annotations produced by other individuals or labs.
}

\dsquestionex{Are there tasks for which the dataset should not be used?}{If so, please provide a description.}

\dsanswer{None.
}

\dsquestion{Any other comments?}

\dsanswer{None.
}

\bigskip
\dssectionheader{Distribution}

\dsquestionex{Will the dataset be distributed to third parties outside of the entity (e.g., company, institution, organization) on behalf of which the dataset was created?}{If so, please provide a description.}

\dsanswer{Yes- the dataset is publicly available for download by all interested third parties.
}

\dsquestionex{How will the dataset will be distributed (e.g., tarball on website, API, GitHub)}{Does the dataset have a digital object identifier (DOI)?}

\dsanswer{The dataset is available on the Caltech public data repository at \href{https://data.caltech.edu/records/1991}{https://data.caltech.edu/records/1991}, where it will be retained indefinitely and available for download by all third parties. The data.caltech.edu posting has accompanying DOI \href{https://doi.org/10.22002/D1.1991}{https://doi.org/10.22002/D1.1991}.

The dataset as used for the MABe Challenge (anonymized sequence and behavior ids) is available for download on the AIcrowd page, located at (\href{https://www.aicrowd.com/challenges/multi-agent-behavior-representation-modeling-measurement-and-applications}{link}). 

}

\dsquestion{When will the dataset be distributed?}

\dsanswer{The full dataset was made publicly available on data.caltech on June 6th, 2021. 
}

\dsquestionex{Will the dataset be distributed under a copyright or other intellectual property (IP) license, and/or under applicable terms of use (ToU)?}{If so, please describe this license and/or ToU, and provide a link or other access point to, or otherwise reproduce, any relevant licensing terms or ToU, as well as any fees associated with these restrictions.}

\dsanswer{The CalMS21 dataset is distributed under the CreativeCommons Attribution-NonCommercial-ShareAlike license (CC-BY-NC-SA). The terms of this license may be found at \href{https://creativecommons.org/licenses/by-nc-sa/2.0/legalcode}{https://creativecommons.org/licenses/by-nc-sa/2.0/legalcode}.
}

\dsquestionex{Have any third parties imposed IP-based or other restrictions on the data associated with the instances?}{If so, please describe these restrictions, and provide a link or other access point to, or otherwise reproduce, any relevant licensing terms, as well as any fees associated with these restrictions.}

\dsanswer{There are no third party restrictions on the data.
}

\dsquestionex{Do any export controls or other regulatory restrictions apply to the dataset or to individual instances?}{If so, please describe these restrictions, and provide a link or other access point to, or otherwise reproduce, any supporting documentation.}

\dsanswer{No export controls or regulatory restrictions apply.
}

\dsquestion{Any other comments?}

\dsanswer{None.
}

\bigskip
\dssectionheader{Maintenance}

\dsquestion{Who will be supporting/hosting/maintaining the dataset?}

\dsanswer{The dataset is hosted on the Caltech Research Data Repository at \href{https://data.caltech.edu/}{data.caltech.edu}. Dataset hosting is maintained by the library of the California Institute of Technology. Long-term support for users of the dataset is provided by Jennifer J. Sun and by the laboratory of Ann Kennedy.
}

\dsquestion{How can the owner/curator/manager of the dataset be contacted (e.g., email address)?}

\dsanswer{The managers of the dataset (JJS and AK) can be contacted at \href{mailto:mabe.workshop@gmail.com}{mabe.workshop@gmail.com}, or AK can be contacted at \href{mailto:ann.kennedy@northwestern.edu}{ann.kennedy@northwestern.edu} and JJS can be contacted at \href{mailto:jjsun@caltech.edu}{jjsun@caltech.edu}.
}

\dsquestionex{Is there an erratum?}{If so, please provide a link or other access point.}

\dsanswer{No.
}

\dsquestionex{Will the dataset be updated (e.g., to correct labeling errors, add new instances, delete instances)?}{If so, please describe how often, by whom, and how updates will be communicated to users (e.g., mailing list, GitHub)?}

\dsanswer{Users of the dataset have the option to subscribe to a mailing list to receive updates regarding corrections or extensions of the CalMS21 dataset. Mailing list sign-up can be found on the CalMS21 webpage at \href{https://sites.google.com/view/computational-behavior/our-datasets/calms21-dataset}{https://sites.google.com/view/computational-behavior/our-datasets/calms21-dataset}.

Updates to correct errors in the dataset will be made promptly, and announced via update messages posted to the CalMS21 website and data.caltech.edu page.

Updates that extend the scope of the dataset, such as additional data sequences, new challenge tasks, or improved pose estimation, will be released as new named instantiations on at most a yearly basis. Previous versions of the dataset will remain online, but obsolescence notes will be sent out to the CalMS21 mailing list. In updates, dataset version will be indicated by the year in the dataset name (here 21). Dataset updates may accompany new instantiations of the MABe Challenge.
}

\dsquestionex{If the dataset relates to people, are there applicable limits on the retention of the data associated with the instances (e.g., were individuals in question told that their data would be retained for a fixed period of time and then deleted)?}{If so, please describe these limits and explain how they will be enforced.}

\dsanswer{N/a (no human data.)
}

\dsquestionex{Will older versions of the dataset continue to be supported/hosted/maintained?}{If so, please describe how. If not, please describe how its obsolescence will be communicated to users.}

\dsanswer{Yes, the dataset will be permanently available on the Caltech Research Data Repository (data.caltech.edu), which is managed by the Caltech Library.
}

\dsquestionex{If others want to extend/augment/build on/contribute to the dataset, is there a mechanism for them to do so?}{If so, please provide a description. Will these contributions be validated/verified? If so, please describe how. If not, why not? Is there a process for communicating/distributing these contributions to other users? If so, please provide a description.}

\dsanswer{Extensions to the dataset will take place through at-most-yearly updates. We welcome community contributions of behavioral data, novel tracking methods, and novel challenge tasks; these may be submitted by contacting the authors or emailing \href{mailto:mabe.workshop@gmail.com}{mabe.workshop@gmail.com}. All community contributions will be visually reviewed by the managers of the dataset for quality of tracking and annotation data; for new challenge tasks, new baseline models will be developed prior to launch to ensure task feasibility. Community contributions will not be accepted without a data maintenance plan (similar to this document), to ensure support for future users of the dataset.
}

\dsquestion{Any other comments?}

\dsanswer{If you enjoyed this dataset and would like to contribute other multi-agent behavioral data for future versions of the dataset or MABe Challenge, contact us at \href{mailto:mabe.workshop@gmail.com}{mabe.workshop@gmail.com}!
}

\section{Data Format}~\label{sec:dataformat}

Our dataset is released in the json format. Each sequence (video) has associated keypoints, keypoint confidence scores, and behavior annotations, all stored as lists in a dictionary, as well as a dictionary of associated metadata, which is nested within the sequence dictionary (see sample below). The unlabeled set is an exception, as it only contains keypoints and scores, with no annotations or metadata fields. For each task, there is one train file and one test file. The train file is used during development and a held out validation set can be used for hyperparameter tuning. The results are reported on the test file.

For all CalMS21 data, the json format is shown in Listing~\ref{listing:1}. Note that the number of frames are the number of frames of each video, so this number could vary across sequences.

\begin{listing}[!htbp]
\begin{minted}[frame=single,
              framesep=3mm,
              linenos=true,
              xleftmargin=21pt,
              tabsize=4]{js}
{   
    "<GROUPNAME>"{
        "<sequence_id-1>": {
            "keypoints" : a list of shape (frames, 2, 2, 7),
            "scores" : a list of shape (frames, 2, 7),
            "annotations" : a list of shape (frames),
            "metadata" : {
                "annotator_id": a number identifying the annotator
                "vocab": a dictionary of behavior names
                }
        },
        "<sequence_id-2>": {
            "keypoints" : a list of shape (frames, 2, 2, 7),
            "scores" : a list of shape (frames, 2, 7),
            "annotations" : a list of shape (frames),
            "metadata" : {
                "annotator_id": a number identifying the annotator
                "vocab": a dictionary of behavior names
        },
        ...
    },
    ...
}
\end{minted}
\caption{Json file format.}\label{listing:1}
\end{listing}

The layer \verb|GROUPNAME| groups together sequences with a similar property, such as a common annotator id. In Task 1, \verb|GROUPNAME| is \verb|annotator_id-0|, and there is only one GROUP in the file. Task 2, \verb|GROUPNAME| is \verb|annotator_id-X|, and there are five groups for $X\in (1,2,3,4,5)$. In Task 3, \verb|GROUPNAME| the name of a behavior.

The \verb|keypoints| field contains the (x,y) position of anatomically defined pose keypoints tracked using MARS\cite{segalin2020mouse}. The dimensions $(2 \times 2 \times 7 )$ correspond to the mouse ID (mouse 0 is the resident and mouse 1 is the intruder), image (x,y) coordinates in pixels, and keypoint ID. For keypoint ID, there are seven tracked body parts, ordered (nose, left ear, right ear, neck, left hip, right hip, tail base).

The \verb|scores| field corresponds to the confidence from the MARS tracker~\cite{segalin2020mouse}, and its dimensions in each frame $(2 \times 7 )$ corresponds to the mouse ID and keypoint ID. 

The \verb|annotations| field contains the frame-level behavior annotations from domain experts as a list of integers.

The \verb|metadata| dictionary for all tasks (except the unlabeled data) contains two fields, an integer \verb|annotator_id| and a dictionary \verb|vocab| which gives the mapping from behavior classes to integer values for the \verb|annotations| list. For example, in Task 1 \verb|vocab| is \verb|attack: 0, investigation: 1, mount: 2, other: 3|.

The dataset website \href{https://sites.google.com/view/computational-behavior/our-datasets/calms21-dataset}{https://sites.google.com/view/computational-behavior/our-datasets/calms21-dataset} also contains a description of the data format and code to load the data for each task.


\section{Dataset Preparation}~\label{sec:datasetpreparation}
\subsection{Behavior Video Acquisition}
This section is adapted from \cite{segalin2020mouse}. Experimental mice ("residents") were transported in their homecage to a behavioral testing room, and acclimatized for 5-15 minutes. Homecages were then inserted into a custom-built hardware setup\cite{hong2015automated} where behaviors are recorded under dim red light condition using a camera (Point Grey Grasshopper3) located 46cm above the homecage floor. Videos are acquired at 30 fps and 1024x570 pixel resolution using StreamPix video software (NorPix). Following two further minutes of acclimatization, an unfamiliar group-housed male or female BALB/c mouse ("intruder") was introduced to the cage, and animals were allowed to freely interact for a period of approximately 10 minutes. BALB/c mice are used as intruders for their white coat color (simplifying identity tracking), as well as their relatively submissive behavior, which reduces the likelihood of intruder-initiated aggression.

\subsection{Behavior Annotation}~\label{sec:behavior}
Behaviors were annotated on a frame-by-frame basis by a trained human expert. Annotators were provided with simultaneous top- and front-view video of interacting mice, and scored every video frame for close investigation, attack, and mounting, defined as follows (reproduced from \cite{segalin2020mouse}):

\begin{enumerate}
\item \textbf{Close investigation}: resident (black) mouse is in close contact with the intruder (white) and is actively sniffing the intruder anywhere on its body or tail. Active sniffing can usually be distinguished from passive orienting behavior by head bobbing/movements of the resident's nose.
\item \textbf{Attack}: high-intensity behavior in which the resident is biting or tussling with the intruder, including periods between bouts of biting/tussling during which the intruder is jumping or running away and the resident is in close pursuit. Pauses during which resident/intruder are facing each other (typically while rearing) but not actively interacting should not be included.
\item \textbf{Mount}: behavior in which the resident is hunched over the intruder, typically from the rear, and grasping the sides of the intruder using forelimbs (easier to see on the Front camera). Early-stage copulation is accompanied by rapid pelvic thrusting, while later-stage copulation (sometimes annotated separately as intromission) has a slower rate of pelvic thrusting with some pausing: for the purpose of this analysis, both behaviors should be counted as mounting, however periods where the resident is climbing on the intruder but not attempting to grasp the intruder or initiate thrusting should not. While most bouts of mounting are female-directed, occasional shorter mounting bouts are observed towards males; this behavior and its neural correlates are described in \cite{karigo2021distinct}.
\end{enumerate}

Annotation was performed either in BENTO\cite{segalin2020mouse} or using a custom Matlab interface. In most videos, the majority of frames will not include one of these three behaviors (see Table \ref{table:task1percentages}): in these instances, animals may be apart from each other exploring other parts of the arena, or may be close together but not actively interacting. These frames are labeled as "other". Because this is not a true behavior, we do not consider classifier performance in predicting "other" frames accurately.
\begin{table}[b]
\begin{center}
 \begin{tabular}{||c | c||} 
 \hline
 Behavior & Percent of Frames \\ [0.5ex] 
 \hline
 attack & 2.76  \\
 \hline
 investigation & 28.9   \\
 \hline
 mount & 5.64\\
 \hline
 other & 62.7   \\
 \hline
\end{tabular}
\end{center}
\caption{The percentage of frames labeled as attack, investigation, mount, and other in the Task 1 training set.}
\label{table:task1percentages}
\end{table}

\subsection{Pose Estimation}
The poses of mice in top-view recordings are estimated using the Mouse Action Recognition System (MARS,\cite{segalin2020mouse}), a computer vision tool that identifies seven anatomically defined keypoints on the body of each mouse: the nose, ears, base of neck, hips, and tail (Figure \ref{fig:keypoint_ids}). MARS estimates animal pose using a stacked hourglass model \cite{newell2016stacked} trained on a dataset of 15,000 video frames, in which all seven keypoints were manually annotated on each of two interacting mice (annotators were instructed to estimate the locations of occluded keypoints.) To improve accuracy, each image in the training set was annotated by five human workers, and "ground truth" keypoint locations were taken to be the median of the five annotators' estimates of each point. All videos in the CalMS21 Dataset were collected in the same experimental apparatus as the MARS training set~\cite{hong2015automated}.

\section{Evaluation}~\label{sec:evaluation}
For all three Tasks, we evaluate performance of trained classifiers in terms of the F1 score and Average Precision for each behavior of interest. Because of the high class imbalance in behavior annotations, we use an unweighted average across behavior classes to compute a single F1 score and Mean Average Precision (MAP) for a given model, omitting the "other" category (observed when a frame is not positive for any annotated behavior) from our metrics.

\paragraph{F1 score.} The F1 score is the harmonic mean of the Precision $P$ and Recall $R$:
\begin{align}
P = \frac{TP}{TP + FP} \\
R = \frac{TP}{TP + FN} \\
F1 = \frac{2 \times P \times R}{P+R}
\end{align}
Where true positives (TP) is the number of frames that a model correctly labels as positive for a class, false positives (FP) is the number of frames incorrectly labeled as positive for a class, and false negatives (FN) is the number of frames incorrectly labeled as negative for a class.

The F1 score is a useful measure of model performance when the number of true negatives (TN, frames correctly labeled as negative for a class) in a task is high. This is the case for the CalMS21 dataset, where for instance attack occurs on less than 3\% of frames.

\paragraph{Average Precision.} The AP approximates the area under the Precision-Recall curve for each behavior class. There are a few different ways to approximate AP; here we compute AP using the implementation from Scikit-Learn~\cite{scikit-learn}:
\begin{align}
AP = \sum_{n} P_n(R_n - R_{n-1})
\end{align}
where $P_n$ and $R_n$ are the precision and recall at the n-th threshold. This implementation is not interpolated. We call the unweighted class-averaged AP the mean average precision (MAP).

\paragraph{Averaging Across Behaviors and Annotators.} Our baseline code, released at \href{https://gitlab.aicrowd.com/aicrowd/research/mab-e/mab-e-baselines}{https://gitlab.aicrowd.com/aicrowd/research/mab-e/mab-e-baselines} shows how we computed our metrics. To compare against our benchmarks, the F1 score and MAP should be computed as follows: for Task 1, the metrics should be computed on the entire test set with all the videos concatenated into one sequence (each frame is weighted equally for each behavior); for Task 2, the metrics should be computed separately on the test set of each annotator, then averaged over the annotators (each behavior and annotator is weighted equally); for Task 3, the metrics should be computed separately for the test set of each behavior, then averaged over the behaviors (each behavior is weighted equally). For our evaluation, the class with the highest predicted probability in each frame was used to compute F1 score, but the F1 score will likely be higher with threshold tuning.

\section{Implementation Details}~\label{sec:implementationdetails}

For more details on the implementation and exact hyperparameter settings, see our code links in Section~\ref{sec:license}.

\subsection{Baseline Model Input}~\label{sec:baseline_input}
Each frame in the CalMS21 dataset is represented by a flattened vector of 28 values, representing the (x,y) location of 7 keypoints from each mouse (resident and intruder). For our baselines, we normalized all (x,y) coordinates by the resolution of the video ($1024 \times 570$ pixels). Associated with these (x,y) values is a single behavior label per frame: in Tasks 1: Classic Classification and Task 2: Annotation Style Transfer, labels may be ``attack", ``mount", ``investigation", or ``other" (i.e. none of the above), while in Task 3: New Behaviors, we provide a separate set of binary labels for each behavior of interest.

We do not require behavior classification models to be causal, so information from both past and future frames can be used for behavior classification. Thus, in the most general form, the input to our model is a stacked set of keypoints from the immediate past, the present frame, and the immediate future, and the model is trained to predict the behavior label only for the present frame (Figure \ref{fig:neuralnetflow}). We refer to our input stack of poses across frames as an input \emph{trajectory}, where the number of past and future frames included in the input trajectory is a model hyperparameter.

Neighboring frames in input trajectories are highly correlated. Therefore, to sample a broad temporal window without significantly increasing the dimensionality of model input, we introduced a skip factor as a second hyperparameter. For a given input trajectory, a skip factor of $N$ signifies that only every $N^{th}$ frame is used when sampling past/future frames. Given current frame $t$, sampling $50$ future frames with a skip factor of $2$ would produce a stack of frames $\{t, t+2, t+4, ... t+(2 \times 50)\}$. We note that more sophisticated compression methods, such as non-uniform downsampling or interpolation, could lead to better representations of temporal data.

We explored model performance as a function of these hyperparameters. For Task 1, we found that models generally performed well when including 100 past frames and 100 future frames, with a skip factor of 2-- ie, model input was every other frame from 200 frames (6.667 seconds) before the current frame to 200 frames after the current frame. For Tasks 2 and 3, we included 50 past frames and 50 future frames with a skip factor of 1. 

\subsection{Baseline Data Augmentation}
Behavior labels should be invariant to certain features of the raw pose data, such as the absolute positions of the agents. We therefore augmented our training data using trajectory transformations, including random rotations, translations, and reflections of the pair of mice. To preserve temporal and relative spatial structure, the same transformation was applied to all frames in a given input trajectory, and to all keypoints from both mice on each frame.

This data augmentation method did not significantly improve model performance for Task 1, although it did improve performance on the more data-limited Tasks 2 and 3. It is possible that a more thorough form of data augmentation, incorporating additional domain-specific knowledge of animal behavior, could further improve model performance. Alternatively, performance on behavior classification tasks could be improved by using domain knowledge to remove non-informative sources of variance, for example by transforming animal trajectories from allocentric to egocentric coordinates.

\subsection{Baseline Task 2 Annotation Style Transfer Details}

The Task 2 baseline model is fine-tuned on the trained model from Task 1. We found that allowing all weights of our pre-trained model to be modified during fine-tuning often resulted in overfitting of the training set, causing a drop in model performance relative to the pre-trained model. This likely arises in part due to class imbalance, which has a more substantial effect on model performance when the training set is small.

To overcome this, we trained the fine-tuned network in phases. In the first phase of fine-tuning, we froze weights in all layers except for the final, densely connected output layer, and trained for a set number of epochs. The number of initial fine-tuning epochs is another model hyperparameter which we tuned for each annotator. Following this initial period of fine-tuning, we unfroze weights in all layers and train until convergence on a validation set. We then used this set of final hyperparameters and trained on the full Task 2 training set. 

\subsection{Baseline Task 3 New Behaviors Details}

As in Task 2, we fine-tuned our model for each behavior in two steps: first freezing all but the output layer for a fixed number of epochs, and then unfreezing all weights and allowing training to run to completion; the learning rate was hand-tuned on a per-behavior basis based on a held-out validation set ($20\%$ of the train split). Once the hyperparameters are chosen, we then trained the final models on the full train split.

To better address the extreme class imbalance in Task 3, we also applied a weight on different behavior samples during training (Table~\ref{table:task3weight}).

\begin{table}
\begin{center}
 \begin{tabular}{|c | c | c |} 
 \hline
 Behavior & \% Frames & Weight  \\ [0.5ex] 
 \hline
approach & 3.25 & 20  \\ \hline
disengaged & 1.47 & 50  \\ \hline
groom & 14.0 & 5  \\ \hline
intromission & 19.3 & 3 \\ \hline
mount attempt & 0.90 & 100  \\ \hline
sniff face & 4.74 & 20  \\ \hline
whiterearing & 9.29 & 10 \\ \hline
\end{tabular}
\end{center}
\caption{Weighting applied on each class for Task 3.}
\label{table:task3weight}
\end{table}

\subsection{Tasks 1 \& 2 Top-1 Entry Details}~\label{sec:task1_implementation_details}

\begin{figure*}
    \centering
    \includegraphics[width=0.8\linewidth]{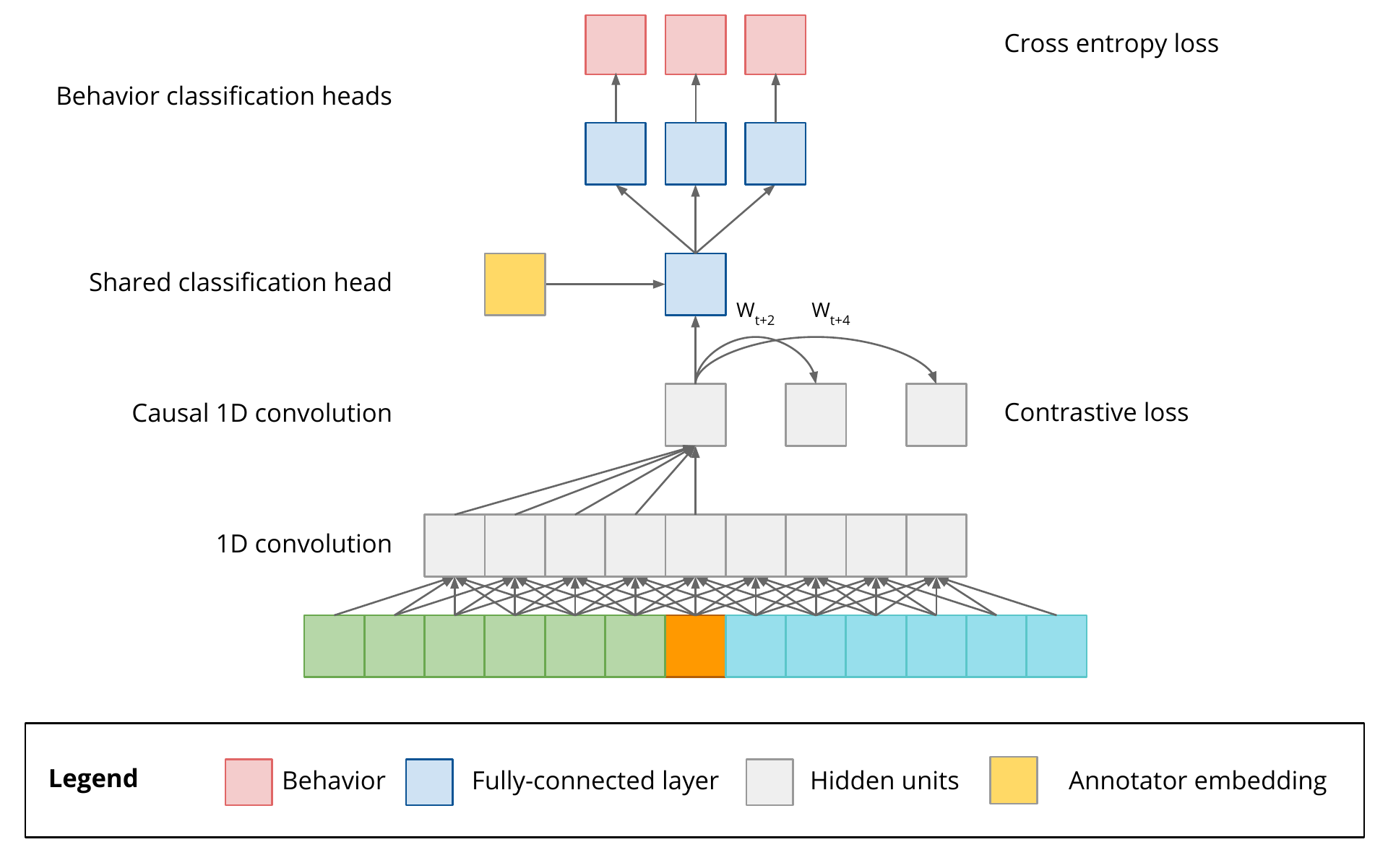}
    \caption{\textbf{Challenge Top-1 Model on Tasks 1 \& 2.}}
    \label{fig:task2_model}
\end{figure*}

The top-1 entry (Figure~\ref{fig:task2_model}) for tasks 1 \& 2 used joint training on all three tasks, including the unlabeled set. The total input dimensionality at each frame is 52. First, 37 features in total were computed based on velocity, egocentric angles, and distances on top of the trajectory data. In addition, 15 PCA dimensions were computed based on distances between keypoints at each frame.

For this model, the MADGRAD optimizer~\cite{defazio2021adaptivity} with cosine learning rate annealing and a small weight decay of 1e-5 was used to fit the model. Label smoothing was applied to the labels to regularize the model and improve calibration~\cite{muller2019does}. Class rescaling weights were used in the categorical cross-entropy loss function because of the unbalanced distribution of behaviors in the data~\cite{king2001logistic}.

Cross-validation was used to optimize the hyperparameters of the model manually. Due to the long training time of the model and the large variance in the validation F1 scores on different splits of the dataset, an exhaustive search in the hyperparameter space was not possible, and most parameters were not tuned at all. Batch size and the number of training epochs were evaluated in the ranges of 16 to 64 and 30 to 60 and set to 32 and 40 for the final model. Dropout at various positions in the network architecture was tested but removed for the final model because it did not improve validation performance. The number of embedder residual blocks was tuned because there is likely a tradeoff between the number of future frames the network can use and the effectiveness of the unsupervised loss. Values in the range from 3 to 16 were evaluated and set to 8 for the final model. The learning rate was evaluated for the values 0.1, 0.01, 0.005, 0.001, 0.0005, 0.0001 and set to 0.001 for the final model. Due to computational constraints, the hyperparameters related to the unsupervised loss (e.g., number of linear projections/timestep into the future) were not tuned.

For the MABe challenge, an ensemble of models was trained on cross-validation splits of the data. These models were then used to bootstrap soft labels for the unlabeled sequences by computing their average predictions. These labels were then used during training of the model, i.e., an ensemble of the model was used to bootstrap training data for the next iteration of the model. This approach was used to further regularize the model by utilizing the implicit knowledge in an ensemble of models trained on the full dataset. For evaluating reproducibility, only one model from the ensemble was selected and trained across 5 runs.

\subsection{Task 3 Top-1 Entry Details}~\label{sec:task3_implementation_details}

\begin{figure*}
    \centering
    \includegraphics[width=0.9\linewidth]{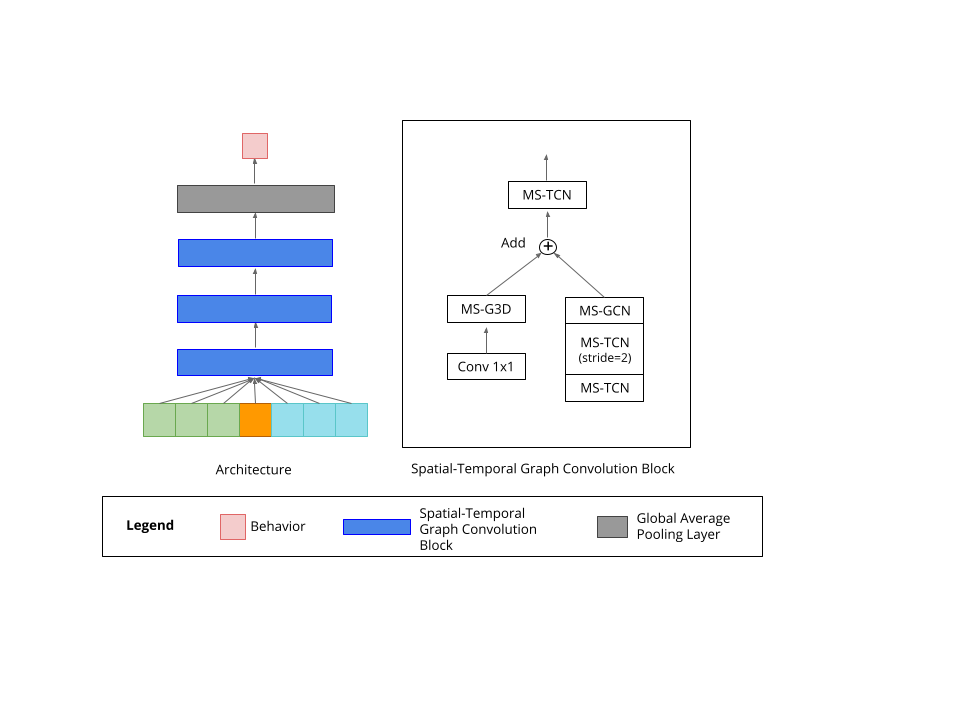}
    \vspace{-0.8in}
    \caption{\textbf{Challenge Top-1 Model on Task 3.} This model follows the MS-G3D Architecture~\cite{liu2020disentangling}. “TCN”, “GCN”, “MS-” respectively denotes temporal and graph convolution blocks (STGC), and multi-scale aggregation.}
    \label{fig:task3_model}
\end{figure*}

For Task 3, the top entry in the MABe challenge trained seven binary classifiers in a fully supervised fashion using the MS-G3D Architecture~\cite{liu2020disentangling} (Figure~\ref{fig:task3_model}). The model starts from keypoints at each frame, as 2D coordinates (X,Y) in pixel coordinate system. The clips are then represented with tensor of (B, 2, T, 14, 1) dimensions, where B and T respectively denotes batch size and sequence length. All models are trained in 10 epochs with SGD with momentum 0.9, batch size 256 (128 per GPU over 2 GPUs), an initial learning rate 0.2 for 6 and 8 epochs with LR weight decay with a factor 0.0003. All skeleton sequences are sampled by 64 frames with stride 2 to have T = 32. Inputs are preprocessed with normlization following~\cite{yan2018spatial}. Random keypoint rotations are used for data augmentation.

During development, since the train split of Task 3 is small, the train split of Task 1 was used during development for validation (a random $20\%$ of the data was held-out). Here, different batch sizes (32/64/128/256), learning rates (0.025/0.05/0.1/0.2), and graph structures (whether the neck keypoints in the spatial graph of the two mice should be connected) was considered. For the final evaluation, all models are trained with the entire train split of task3. 

For the MABe challenge, the winning model consisted of an ensemble of models trained in this way. For evaluating reproducibility, only one model from the ensemble was selected and trained across 5 runs.

\section{Additional Results}~\label{sec:additionalresults}
\subsection{Additional Results for Task 1: Classic Classification}

\begin{table}[t]
    \centering
    \scalebox{0.85}{
    \begin{tabular}{l|cc|cc|cc}
        \toprule[0.2em]
        \multirow{2}{*}{Method} & \multicolumn{2}{c|}{Attack} & \multicolumn{2}{c|}{Investigation} & \multicolumn{2}{c}{Mount} \\
        & F1 & AP & F1 & AP & F1 & AP \\
        \toprule[0.2em]
        Baseline & $.664 \pm .031$ & $.724 \pm .023$ & $.814 \pm .005$ & $.893 \pm .005$ & $.900 \pm .004$ & $.950 \pm .004$ \\
        Baseline w/ task prog & $.789 \pm .002$ & $.839 \pm .011$ & $.817 \pm .003$ & $.889 \pm .006$ & $.880 \pm .011$ & $.939 \pm .009$\\
        MABe 2021 Task 1 Top-1 & $.827 \pm .024$ & $.885 \pm .012$ &  $.852 \pm .011$ & $.908 \pm .014$ &  $.913 \pm .019$ & $.950 \pm .014$ \\
        \bottomrule[0.1em]
    \end{tabular}
    }
    \caption{Per-class results on Task 1 (mean $\pm$ standard deviation over 5 runs.) }\label{table:task1metrics_perclass}   
    \vspace{-0.1in}
\end{table}

Results for each class on Task 1 is in Table~\ref{table:task1metrics_perclass}. The behavior class ``attack" is where we observe the most performance difference on our benchmarks, while model performances on investigation and mount are closer (often within one standard deviation). The performance on the ``mount" class is very close to the ceiling, while both ``attack" and ``investigation" can likely be further improved.

\subsection{Additional Results for Task 2: Annotation Style Transfer}

Depending on the annotator and behavior, models can have a range of performances from lower F1 scores such as $0.6$ to higher ones such as $0.9$ (Table~\ref{table:task2metrics_perclass}). We observe that all classifiers generally performs better on annotators 1 and 2, while performing worse on annotators 3, 4, and 5. This demonstrates the importance of studying techniques that can transfer to different annotation styles. Models that can more accurately classify behavior for all annotators could help annotators better understand the difference in their styles, and could potentially be applicable to more users.

\begin{table}[t]
    \centering
    \scalebox{0.85}{
    \begin{tabular}{l|cc|cc|cc}
        \toprule[0.2em]
        \multirow{2}{*}{Method} & \multicolumn{2}{c|}{Attack} & \multicolumn{2}{c|}{Investigation} & \multicolumn{2}{c}{Mount} \\
        & F1 & AP & F1 & AP & F1 & AP \\
        \toprule[0.2em]
        & \multicolumn{6}{c}{Annotator 1}\\
        \hline 
        Baseline & $.713 \pm .027$ & $.819 \pm .017$ & $.804 \pm .010$ & $.878 \pm .008$ & $.889 \pm .012$ & $.969 \pm .002$ \\
        Baseline w/ task prog & $\mathbf{.868 \pm .001}$ & $\mathbf{.944\pm .001}$ & $.830\pm .001$ & $.906\pm .001$ & $.912\pm .001$ & $\mathbf{.974\pm .001}$\\
        MABe 2021 Task 2 Top-1 & $.858 \pm .012$ & $.893 \pm .028$ & $\mathbf{.858 \pm .012}$ & $\mathbf{.916 \pm .012}$ & $\mathbf{.939 \pm .007}$ & $.968 \pm .005$ \\
        \hline 
        & \multicolumn{6}{c}{Annotator 2}\\
        \hline 
        Baseline & $.839 \pm .008$ & $.887 \pm .006$ & $.865 \pm .006$ & $.924 \pm .006$ & $.671 \pm .050$ & $.836 \pm .019$ \\
        Baseline w/ task prog & $\mathbf{.881 \pm .001}$ & $\mathbf{.938 \pm .001}$ & $.867 \pm .002$ & $.922 \pm .001$ & $\mathbf{.866 \pm .001}$ & $\mathbf{.943 \pm .003}$\\
        MABe 2021 Task 2 Top-1 & $.865 \pm .019$ & $.937 \pm .020$ & $\mathbf{.904 \pm .010}$ & $\mathbf{.954 \pm .004}$ & $.849 \pm .041$ & $.914 \pm .016$ \\ 
        \hline 
        & \multicolumn{6}{c}{Annotator 3}\\
        \hline 
        Baseline & $.687 \pm .022$ & $.710 \pm .013$ & $\mathbf{.637 \pm .006}$ & $.650 \pm .009$ & $\mathbf{.812 \pm .006}$ & $.820 \pm .012$ \\
        Baseline w/ task prog & $\mathbf{.710 \pm .004}$ & $\mathbf{.755 \pm .002}$ & $.583 \pm .002$ & $.637 \pm .001$ & $.773 \pm .004$ & $.866 \pm .003$ \\
        MABe 2021 Task 2 Top-1 & $.707 \pm .032$ & $.746 \pm .013$ & $.635 \pm .020$ & $\mathbf{.660 \pm .011}$ & $.763 \pm .099$ & $\mathbf{.893 \pm .031}$ \\    
        \hline 
        & \multicolumn{6}{c}{Annotator 4}\\
        \hline 
        Baseline & $.736 \pm .012$ & $.721 \pm .017$ & $.686 \pm .010$ & $.732 \pm .011$ & $.882 \pm .032$ & $.947 \pm .009$ \\
        Baseline w/ task prog & $.747 \pm .001$ & $.776 \pm .002$ & $.685 \pm .004$ & $.751 \pm .005$ & $.877 \pm .006$ & $.939 \pm .003$   \\
        MABe 2021 Task 2 Top-1 & $\mathbf{.795 \pm .018}$ & $\mathbf{.793 \pm .031}$ & $\mathbf{.733 \pm .005}$ & $\mathbf{.784 \pm .018}$ & $\mathbf{.895 \pm .068}$ & $\mathbf{.978 \pm .017}$  \\    
        \hline 
        & \multicolumn{6}{c}{Annotator 5}\\
        \hline 
        Baseline & $.624 \pm .023$ & $.715 \pm .010$ & $.818 \pm .004$ & $\mathbf{.880 \pm .006}$ & $.646 \pm .057$ & $.709 \pm .061$ \\
        Baseline w/ task prog & $\mathbf{.732 \pm .002}$ & $\mathbf{.801 \pm .001}$ & $.800 \pm .002$ & $.866 \pm .001$ & $.719  \pm .020$ & $.804  \pm .006$ \\
        MABe 2021 Task 2 Top-1 & $.698 \pm .051$ & $.693 \pm .063$ & $\mathbf{.838 \pm .010}$ & $.874 \pm .024$ & $\mathbf{.791 \pm .060}$ & $\mathbf{.848 \pm .039}$ \\ 
        \bottomrule[0.1em]
    \end{tabular}
    }
    \caption{Per-annotator and per-class results on Task 2 (mean $\pm$ standard deviation over 5 runs. The top performing method is in bold.) }\label{table:task2metrics_perclass}   
    \vspace{-0.1in}
\end{table}

\begin{figure*}
    \centering
    \includegraphics[width=0.6\linewidth]{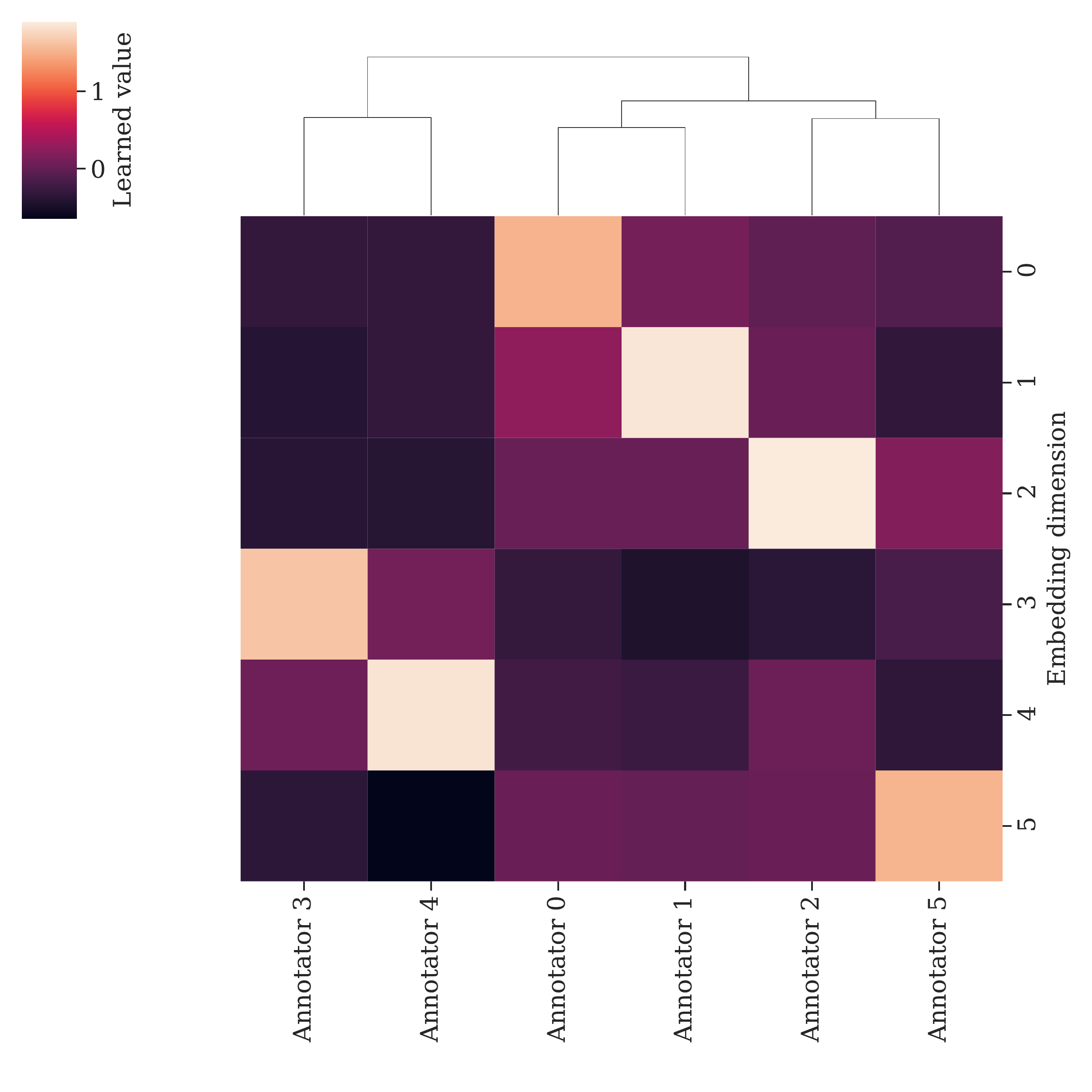}
    \caption{Learned Annotator Matrix from Top-1 Model on Tasks 1 \& 2. The matrix is initialized as a diagonal matrix and each embedding dimension corresponds to the correponding annotator id. The lines on the top of the matrix represent the results from hierarchical clustering.}
    \label{fig:annotator_matrix}
\end{figure*}

\textbf{Annotator Embedding Matrix.} After training the top entry for the Task 1 \& 2 model, the learned annotator embedding matrix was extracted (Figure~\ref{fig:annotator_matrix}). Euclidean distances were computed for all pairs of annotator embeddings and hierarchical clustering was performed using these distances and Ward's method~\cite{ward1963hierarchical}. This learned embedding matrix corresponds to learned similarity between the annotators. Here, we see that annotators 3 and 4 are more different from the other annotators, and are more similar to each other.

\subsection{Additional Results for Task 3: New Behaviors}

Based on performance metrics, Task 3 is more difficult than the other two Tasks (Table~\ref{tab:task3metrics_perclass}). The challenge of this task is the limited amount of training data for each new behavior, and that some behaviors are very rare (for example, mount attempt occurs in less than $1\%$ of the frames). Here, none of the evaluated models consistently perform well on all the behaviors. We note that threshold tuning can help improve F1 score instead of simply taking the class with max predicted probabilities.

\begin{table}[t]
    \centering
    \scalebox{0.85}{
    \begin{tabular}{l|cc|cc|cc}
        \toprule[0.2em]
        \multirow{2}{*}{Method} & \multicolumn{2}{c|}{Approach} & \multicolumn{2}{c|}{Disengaged} & \multicolumn{2}{c}{Groom} \\
        & F1 & AP & F1 & AP & F1 & AP \\
        \toprule[0.2em]
        Baseline & $.338 \pm .011$ & $.282 \pm .005$ & $.195 \pm .005$ & $.237 \pm .017$ & $.289 \pm .023$ & $.260 \pm .028$ \\
        Baseline w/ task prog & $.310 \pm .011$ & $.233 \pm .015$ & $.198 \pm .010$ & $.160 \pm .024$ & $.409 \pm .008$ & $.404 \pm .010$  \\
        \multirow{2}{*}{MABe 2021 Task 3 Top-1} & $.182 \pm .039$ & \multirow{2}{*}{$.209 \pm .016$} & $.101 \pm .033$ & \multirow{2}{*}{$.086 \pm .010$}& $.411 \pm .087$ & \multirow{2}{*}{$.391 \pm .087$}\\
        & ($.272 \pm .031$) & & ($.175 \pm .022$) & & ($.415 \pm .112$)\\
        \bottomrule[0.1em]
        \multirow{2}{*}{Method} & \multicolumn{2}{c|}{Intromission} & \multicolumn{2}{c|}{Mount Attempt} & \multicolumn{2}{c}{Sniff Face} \\
        & F1 & AP & F1 & AP & F1 & AP \\
        \toprule[0.2em]
        Baseline & $.721 \pm .010$ & $.746 \pm .025$ & $.034 \pm .003$ & $.015 \pm .001$ & $.358 \pm .006$ & $.322 \pm .015$ \\
        Baseline w/ task prog & $.609 \pm .019$ & $.698 \pm .023$ & $.048 \pm .009$ & $.021 \pm .004$ & $.274 \pm .026$ & $.314 \pm .022$\\
        \multirow{2}{*}{MABe 2021 Task 3 Top-1} & $.663 \pm .033$ & \multirow{2}{*}{$.761 \pm .021$} & $.001 \pm .002$ & \multirow{2}{*}{$.013 \pm .007$}& $.304 \pm .029$ & \multirow{2}{*}{$.311 \pm .015$}\\
        & ($.697 \pm .026$) & & ($.012 \pm .011$) & & ($.361 \pm .023$)\\
        \bottomrule[0.1em]    
        \multirow{2}{*}{Method} & \multicolumn{2}{c|}{White Rearing}  \\
        & F1 & AP \\
        \toprule[0.2em]
        Baseline & $.430 \pm .006$ & $.355 \pm .017$ & \\
        Baseline w/ task prog & $.430 \pm .029$ & $.427 \pm .025$ & \\
         \multirow{2}{*}{MABe 2021 Task 3 Top-1} & $.569 \pm .035$ & \multirow{2}{*}{$.699 \pm .031$} & \\
        & ($.606 \pm .029$) & \\
        \bottomrule[0.1em]            
    \end{tabular}
    }
    \caption{Per-class results on the seven behaviors in Task 3 (mean $\pm$ standard deviation over 5 runs). The average F1 score in brackets corresponds to improvements with threshold tuning. }\label{tab:task3metrics_perclass}   
    \vspace{-0.1in}
\end{table}

\section{Reproducibility Checklist}~\label{sec:benchmarkreproducibility}


Here we provide additional details based on the ML Reproducibility Checklist.

\subsection{Baselines}
\begin{itemize}
    \item \textbf{Source code link}: \url{https://gitlab.aicrowd.com/aicrowd/research/mab-e/mab-e-baselines}
    \item \textbf{Data used for training}: Train split of the corresponding task. 
    \item \textbf{Pre-processing}: See Sections~\ref{sec:baseline_architectures}, ~\ref{sec:baseline_input}.
    \item \textbf{How samples were allocated for train/val/test}: CalMS21 provides train and test splits. The val split was held out from a random $20\%$ of the train split.
    \item \textbf{Hyperparameter considerations}: See Section~\ref{sec:baseline_architectures}. In particular, for Task 1, we considered learning rates (0.0001/0.0005/0.005/0.001), frame skip (1/2), window size (50, 100, 200), convolution size (3/5/7/9) for the 1D Conv Net, and channel sizes for each layer (16/32/64/128/256). The hyperparameters for Tasks 2 and 3 are based on the tuned hyperparameters on Task 1.
    \item \textbf{Number of evaluation runs}: 5
    \item \textbf{How experiments where ran}: See Section~\ref{sec:baseline_architectures}.
    \item \textbf{Evaluation metrics}: Average F1 and MAP
    \item \textbf{Results}: See Sections~\ref{sec:task1_results}, ~\ref{sec:task2_results},
    ~\ref{sec:task3_results},
    ~\ref{sec:additionalresults}.
    \item \textbf{Computing infrastructure used}: All baseline experiments were ran on Google Colab on CPU (Intel 2.3 GHz Xeon CPU).
\end{itemize}

\subsection{Baselines with Task Programming Features}
\begin{itemize}
    \item \textbf{Source code link}: \url{https://github.com/neuroethology/TREBA}
    \item \textbf{Data used for training}: Unlabeled set for pre-training the features extraction model with task programming. Train split of the corresponding task for training baseline. 
    \item \textbf{Pre-processing}: Same as baselines (see Sections~\ref{sec:baseline_architectures}, ~\ref{sec:baseline_input}) except with task programming features concatenated at each frame.
    \item \textbf{How samples were allocated for train/val/test}: Task programming model was trained on the first 220 sequences of the unlabeled set and validated on last 62 sequences.
    \item \textbf{Hyperparameter considerations}: The task programming model uses the same hyperparameters as~\cite{sun2020task}, except trained for 300 epochs since the training set is smaller. The same hyperparameters as baselines above are used for training the supervised models, except Task 2 requires less epochs to converge (6 epochs instead of 10). 
    \item \textbf{Number of evaluation runs}: 5
    \item \textbf{How experiments where ran}: See Section~\ref{sec:baseline_architectures} except with task programming features concatenated at each frame.
    \item \textbf{Evaluation metrics}: Average F1 and MAP
    \item \textbf{Results}: See Sections~\ref{sec:task1_results}, ~\ref{sec:task2_results},
    ~\ref{sec:task3_results},
    ~\ref{sec:additionalresults}.
    \item \textbf{Computing infrastructure used}: The task programming models are trained on a Amazon p2 instance, with one NVIDIA K80 GPU, and Intel 2.3 GHz Xeon CPU.
\end{itemize}

\subsection{MABe Challenge Tasks 1 \& 2 Top-1}
\begin{itemize}
    \item \textbf{Data used for training}: All splits of CalMS21, including unlabeled videos. 
    \item \textbf{Pre-processing}: See Sections~\ref{sec:task1_results}, ~\ref{sec:task1_implementation_details}.
    \item \textbf{How samples were allocated for train/val/test}: CalMS21 provides train and test splits. See Section~~\ref{sec:task1_implementation_details} for validation set details.
    \item \textbf{Hyperparameter considerations}: See Section~\ref{sec:task1_implementation_details}.
    \item \textbf{Number of evaluation runs}: 5
    \item \textbf{How experiments where ran}: See Section~\ref{sec:task1_implementation_details}.
    \item \textbf{Evaluation metrics}: Average F1 and MAP
    \item \textbf{Results}: See Sections~\ref{sec:task1_results}, ~\ref{sec:task2_results}, ~\ref{sec:additionalresults}.
    \item \textbf{Computing infrastructure used}: A server with 4 x GeForce RTX 2080. Each model for the cross-validation runs was trained on a single GPU. The server has a AMD Ryzen Threadripper 1950X CPU with 64GB RAM.
\end{itemize}

\subsection{MABe Challenge Tasks 3 Top-1}
\begin{itemize}
    \item \textbf{Data used for training}: Train split of Task 3. 
    \item \textbf{Pre-processing}: See Sections ~\ref{sec:task3_results}, ~\ref{sec:task3_implementation_details}.
    \item \textbf{How samples were allocated for train/val/test}: CalMS21 provides train and test splits. The model was validated on $20\%$ of the train split of Task 1.
    \item \textbf{Hyperparameter considerations}: See Section~\ref{sec:task3_implementation_details}.
    \item \textbf{Number of evaluation runs}: 5
    \item \textbf{How experiments where ran}: See Section~\ref{sec:task3_implementation_details}.
    \item \textbf{Evaluation metrics}: Average F1 and MAP
    \item \textbf{Results}: See Sections~\ref{sec:task3_results}, ~\ref{sec:additionalresults}.
    \item \textbf{Computing infrastructure used}: All models are trained on 2 NVIDIA TESLA V100 32GB GPUs.
\end{itemize}

\end{document}